\documentclass{article}

    \PassOptionsToPackage{numbers, compress}{natbib}

 \usepackage[main, final]{neurips_2026}

\usepackage[utf8]{inputenc} 
\usepackage[T1]{fontenc}    
\usepackage{hyperref}       
\usepackage{url}            
\usepackage{booktabs}       
\usepackage{amsfonts}       
\usepackage{nicefrac}       
\usepackage{microtype}      
\usepackage{xcolor}         

\title{To Call or Not to Call: Diagnosing Intrinsic Over-Calling Bias in LLM Agents}

\usepackage{lipsum}
\usepackage{fancyhdr}       
\usepackage{float}
\usepackage{algorithm}
\usepackage{algpseudocode}
\usepackage{multirow}
\usepackage{placeins}
\usepackage{enumitem}
\usepackage{arydshln}       
\usepackage{verbatim}
\usepackage{listings}
\usepackage{caption}
\usepackage{wrapfig}
\usepackage[most]{tcolorbox}
\tcbuselibrary{breakable, skins}
\definecolor{myorange}{RGB}{2, 142, 2}
                                                                                   
\hypersetup{
    colorlinks=true, 
    linkcolor=blue, 
    filecolor=magenta, 
    urlcolor=cyan, 
    citecolor=blue 
}

\newcommand{\ie}{\emph{i.e., }}

\newtcolorbox{prompt}[1]{
    enhanced,
    colback=gray!20,
    colframe=black,
    boxrule=0.3pt,
    arc=3mm,
    left=2pt,
    right=2pt,
    boxsep=3pt,
    fonttitle=\small\bfseries,
    title=#1,
    fontupper=\scriptsize
}
                                                                                      
\newtcolorbox{promptbox}[1][]{                                  
breakable,                                                                                                                
enhanced,                                  
colback=gray!6,                                               
colframe=gray!40,                                                                                                         
boxrule=0.5pt,
arc=3pt,                                                                                                                  
left=6pt, right=6pt, top=4pt, bottom=4pt,
fontupper=\small\ttfamily,                                                                                                
title=#1,                                                                                                                 
fonttitle=\small\bfseries\sffamily,                                                                                       
coltitle=black,                                                                                                           
attach boxed title to top left={yshift=-2mm, xshift=4mm},     
boxed title style={colback=gray!15, colframe=gray!40, boxrule=0.5pt, arc=2pt},                                            
}

\author{
    \textbf{Wei Shi\textsuperscript{1, 2}}\thanks{Equal Contribution},
    \textbf{Ziheng Peng\textsuperscript{2, 3}}\footnotemark[1],
    \textbf{Sihang Li\textsuperscript{5}},
    \textbf{Xiting Wang\textsuperscript{3}},
\\
    \textbf{Xiang Wang\textsuperscript{5}},
    \textbf{Mengnan Du\textsuperscript{4}},
    \textbf{Na Zou\textsuperscript{2}}\thanks{Corresponding}, 
\\
\\
    \textsuperscript{1}Shanghai Jiao Tong University,
    \textsuperscript{2}Shanghai Artificial Intelligence Laboratory,
\\
    \textsuperscript{3}Renmin University of China,
    \textsuperscript{4}The Chinese University of Hong Kong Shenzhen,
\\
    \textsuperscript{5}University of Science and Technology of China,
\\
    \texttt{swei2001@sjtu.edu.cn}, \texttt{mengnandu@cuhk.edu.cn}, \\ 
    \texttt{\{ziheng.peng, xitingwang\}@ruc.edu.cn}, \\
    \texttt{\{sihang0520, xiangwang1223, zouna891252\}@gmail.com} 
}

\begin{document}
\maketitle

\begin{abstract}
\label{sec:abstract}

LLM agents exhibit a consistent tendency to over-call, invoking tools even in situations where none is needed.
On the When2Call benchmark, six models from three families show high call accuracy but much lower no-call accuracy, leaving overall accuracy in the 55\%--70\% range.
We trace this to an Intrinsic Bias Hypothesis (\textbf{IBH}): the call/no-call decision mapping carries an activation-independent \textsc{call} offset, so the model favors \textsc{call} even at activation parity.
Using Sparse Autoencoders (SAEs), we recover behavior-aligned feature bases for the \textsc{call}/\textsc{no\_call} decision, reduce them to a signed activation margin, and estimate the offset directly.
Across all six models, the model is decision-neutral only when \textsc{no\_call} activation outweighs \textsc{call} activation, consistent with IBH.
We then causally test IBH with Adaptive Margin-Calibrated Steering (\textbf{AMCS}), a closed-form counter-bias shift along SAE decoder directions.
Cancelling the diagnosed offset mitigates over-calling and improves overall accuracy with a negligible drop in call accuracy.
Our work recasts over-calling from an empirical phenomenon into a mechanistic object amenable to causal correction. 
The code is available at \url{https://github.com/SKURA502/agent-sae}.

\end{abstract}

\section{Introduction}
\label{sec:introduction}

Large language models (LLMs) have rapidly evolved from text generators into the reasoning backbone of autonomous agents~\cite{gpt5,opus46,gemma4,qwen35}.
At the core of this transition lies tool use, the ability to interface with external systems such as search engines, code interpreters, APIs, and databases, extending LLMs beyond pure language modeling~\cite{toolformer,toolllm,gorilla,aflow}.

\begin{figure}[t]
  \centering
  \includegraphics[width=\linewidth]{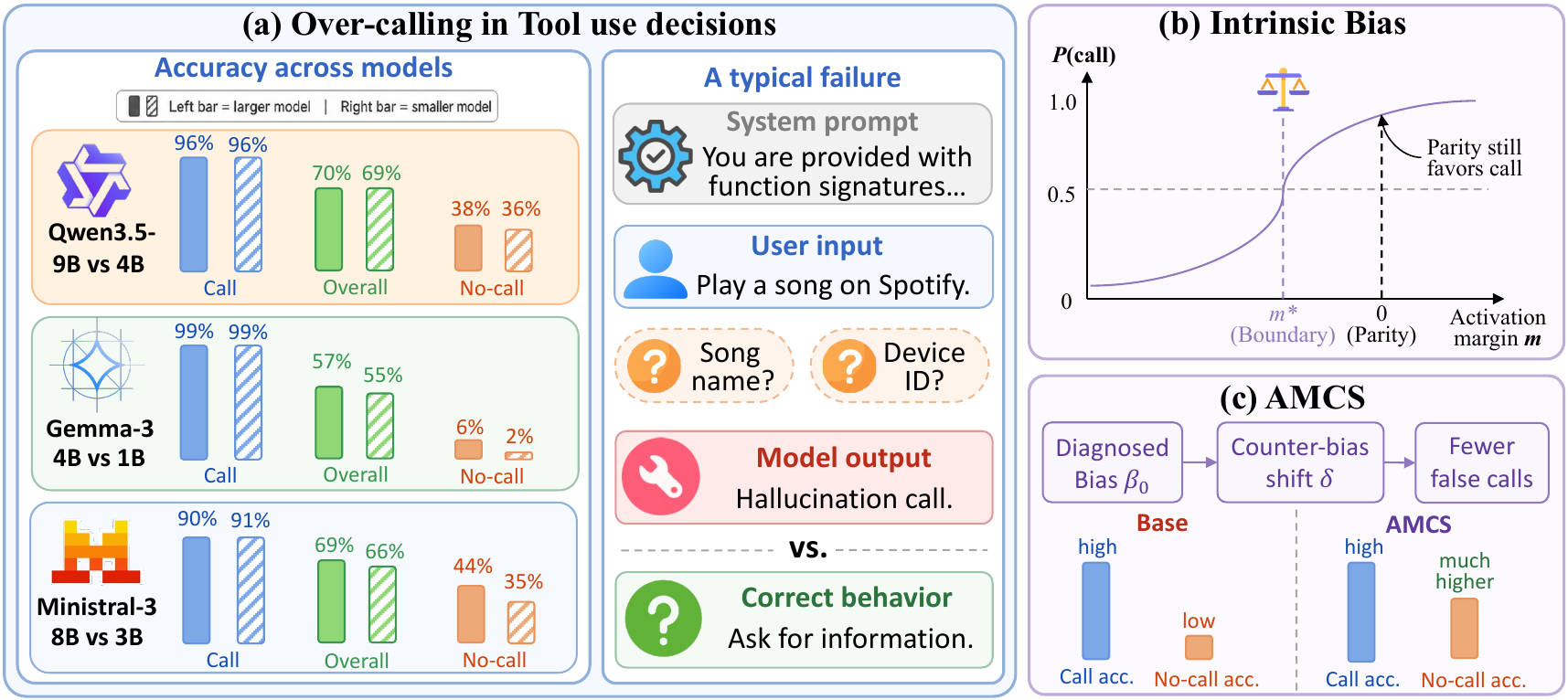}
  \caption{\textbf{Overview of over-calling, intrinsic bias, and AMCS.}
    \textbf{(a)}~Across six target models, \textcolor{blue}{call accuracy} is high, but \textcolor{orange}{no-call accuracy} is much lower, reducing \textcolor{green!60!black}{overall accuracy}; a representative case shows the model calling a tool despite missing required information.
    \textbf{(b)}~Intrinsic Bias Hypothesis: at activation parity ($m=0$), the decision still favors \textsc{call}; the neutral boundary shifts to $m^\star<0$.
    \textbf{(c)}~AMCS converts the diagnosed bias $\beta_0$ into a counter-bias shift $\delta$, aiming to reduce false calls while preserving valid calls.
  }
  \label{fig:teaser}
  \vspace{-14pt}
\end{figure}

Beyond executing calls correctly, effective tool use hinges on knowing when to invoke a tool and, crucially, when not to.
As shown in Figure~\ref{fig:teaser}(a), across six models from three families (Qwen3.5~\cite{qwen35}, Gemma-3~\cite{gemma3}, Ministral-3~\cite{ministral3}) evaluated on the When2Call benchmark~\cite{when2call}, call accuracy remains high, while no-call accuracy is consistently much lower, leaving overall accuracy in the 55\%--70\% range.
Models therefore know how to call tools when calls are required, but often issue calls when none is warranted.
Figure~\ref{fig:teaser}(a) illustrates this failure mode: given an underspecified Spotify request, the base model invokes the tool instead of asking for the missing song name and device ID.
This bias degrades user experience and inflates API costs in deployed systems.

Having documented the bias, we ask what governs the call/no-call decision.
An activation-only account offers a natural mechanism: the decision is determined by the relative activation of call- and no-call-related directions, with the stronger side prevailing.
This account makes a testable prediction.
If the decision depends only on the activation difference, balancing call and no-call activation should leave the model decision-neutral.
Our analysis shows the opposite.
Even at activation parity, the model remains biased toward \textsc{call}; it becomes neutral only when no-call activation is stronger.
A residual bias at parity cannot come from activation levels themselves and must instead enter the decision through a separate, additive term.
We formalize this as the Intrinsic Bias Hypothesis (\textbf{IBH}): over-calling reflects an activation-independent call offset in the decision mapping.

We test IBH through a pipeline built on Sparse Autoencoders (SAEs)~\cite{vanillaSAE,topkSAE}, which decompose residual-stream activations into sparse and interpretable features~\cite{vanillaSAE,claudeScaling} and let us isolate the components driving the \textsc{call}/\textsc{no\_call} decision.
We first recover behavior-aligned \textsc{call} and \textsc{no\_call} feature bases and verify that a handful of them predict the decision near the residual-stream upper bound (\S\ref{sec:discovery}).
We then reduce these bases to a signed activation margin and test whether activation parity removes the decision asymmetry (\S\ref{sec:diagnosis}).
It does not: activation geometry and logistic estimates of the offset both show a shifted neutral boundary (Figure~\ref{fig:teaser}(b)), and the same SAE directions also separate true from false calls, exposing a lever for intervention.
Finally, we causally test IBH with Adaptive Margin-Calibrated Steering (\textbf{AMCS}), a closed-form counter-bias shift along SAE decoder directions (\S\ref{sec:steering}; Figure~\ref{fig:teaser}(c)).
Cancelling the diagnosed offset raises no-call accuracy by 4--17 points on five of six models and overall accuracy by up to 5 points, with minimal impact on call accuracy across the same models.

Our main contributions are:
\begin{itemize}[leftmargin=2em]
  \item We open a mechanistic view of tool-use gating by recovering behavior-aligned SAE feature bases that predict the \textsc{call}/\textsc{no\_call} decision near the residual-stream upper bound (\S\ref{sec:discovery}).
  \item We formulate and validate \textbf{IBH}, showing that over-calling reflects an activation-independent call offset rather than activation levels alone, and that the same SAE directions further separate true from false calls (\S\ref{sec:diagnosis}).
  \item We introduce \textbf{AMCS}, a closed-form steering method derived from the diagnosed offset, and use it as a causal validation of \textbf{IBH} that mitigates over-calling while preserving valid tool calls (\S\ref{sec:steering}).
\end{itemize}

\section{Related Work}
\label{sec:related_work}

\subsection{Tool-Use Evaluation and the Over-Calling Phenomenon}
LLM agents have evolved from early API and calculator integrations~\cite{toolformer,mrkl,react} to systems spanning large collections of real-world tools~\cite{toolllm}.
A corresponding evaluation ecosystem studies whether models can invoke tools correctly: selecting the right function, producing valid arguments, and completing tool-mediated tasks~\cite{gorilla,agentbench,bfcl}.
These benchmarks measure how well models call tools, but the gating question of when not to call is less central.
When2Call~\cite{when2call} targets this decision directly and reveals a consistent asymmetry: models perform much better on call-required queries than on queries where not calling a tool is correct.
Rather than another benchmark or data-side remedy, we ask what internal mechanism produces this asymmetry and diagnose over-calling as a measurable bias in the model's decision mapping.

\subsection{Mechanistic Interpretability and Sparse Autoencoders}
Understanding why over-calling arises requires tools that can dissect a model's internal computations.
Mechanistic interpretability~\cite{openProblems} provides this foundation, aiming to decompose model behavior into interpretable internal components and identify the structures responsible for specific outputs.
Within this framework, Sparse Autoencoders (SAEs) have become a widely adopted method for recovering sparse, interpretable units from residual streams~\cite{vanillaSAE, claudeTowards, claudeScaling, sarm, Concept2Scenario}, with variants such as TopK~\cite{topkSAE}, JumpReLU~\cite{jumpreluSAE}, and RouteSAE~\cite{routeSAE} improving reconstruction fidelity and feature quality.

We use SAEs to extract a structured feature basis for testing a mechanistic hypothesis.
The same basis supports both estimating an activation-independent call offset and applying a counter-bias intervention along these directions as a causal test of the diagnosed offset.

\section{Discovering Gating Feature Bases}
\label{sec:discovery}

Testing IBH requires a feature basis that reliably indexes the model's call/no-call decision.
We construct two such bases, $\mathcal{C}$ and $\mathcal{N}$, by training SAEs on residual streams (\S\ref{sec:sae_training}), ranking features by the model's observed call/no-call behavior (\S\ref{sec:discovery_pipeline}), and validating them with linear probes (\S\ref{sec:validation}).

\subsection{SAE Training}
\label{sec:sae_training}

We train a separate TopK SAE~\cite{topkSAE} for each of six target models from three families at two scales: Qwen3.5-(4B, 9B), Gemma-3-it-(1B, 4B), and Ministral-3-Instruct-(3B, 8B).
Given a residual-stream activation $\mathbf{h} \in \mathbb{R}^d$, the encoder and decoder compute:
\begin{align}
  \mathbf{z} &= \mathrm{TopK}\!\left(\mathbf{W}_{\mathrm{enc}}(\mathbf{h} - \mathbf{b}_{\mathrm{pre}})\right) \;\in\; \mathbb{R}^{M}, \label{equ:sae_encoder} \\
  \hat{\mathbf{h}} &= \mathbf{W}_{\mathrm{dec}}\,\mathbf{z} + \mathbf{b}_{\mathrm{pre}}, \label{equ:sae_decoder}
\end{align}
where $M = 8d$, $K = \lfloor d/32 \rfloor$, and the columns of $\mathbf{W}_{\mathrm{dec}} \in \mathbb{R}^{d \times M}$ are constrained to unit norm.
Training minimizes the reconstruction loss $\mathcal{L} = \|\mathbf{h} - \hat{\mathbf{h}}\|_{2}^{2}$.
Each SAE is hooked at the output residual stream of a middle-to-late transformer block of its target model, where representations are sufficiently abstract for high-level decision features to emerge.
Training follows a two-stage curriculum.
Stage 1 trains on OpenWebText2~\cite{pile} (${\approx}50$M tokens) to learn a broad sparse feature basis grounded in general residual-stream geometry.
We use this broad-corpus stage to reduce the risk that a narrow, domain-specific initialization leaves gaps in residual-stream coverage.
Stage 2 continues on the When2Call training split (${\approx}10$M tokens) to adapt the dictionary to tool-use contexts while retaining the broad coverage from Stage 1.
Training details, model-specific hook locations, and diagnostics are reported in Appendix~\ref{app:sae_training}.

\subsection{Behavior-Labeled Feature Ranking}
\label{sec:discovery_pipeline}

\paragraph{Dataset and behavioral labeling.}
We build the discovery set from the When2Call evaluation split~\cite{when2call}, using all contexts regardless of the original category.
For each context $x_i$, we run the target LLM $f_\theta$ to obtain a response $y_i$ and use an independent LLM judge to classify $y_i$ into one of the four When2Call response types (judge prompt in Appendix~\ref{app:judge_prompt}).
Responses judged as \texttt{tool\_call} form $\mathcal{D}^{+}$ (\textsc{call}) and responses judged as \texttt{request\_for\_info} form $\mathcal{D}^{-}$ (\textsc{no\_call}). Responses in the other two categories are excluded.
Each $(x_i, d_i)$ records the model's observed gating decision rather than an external correctness label.
Per-model response-label counts are reported in Appendix~\ref{app:behavior_counts}.

\begin{figure}[h]
  \centering
  \includegraphics[width=\linewidth]{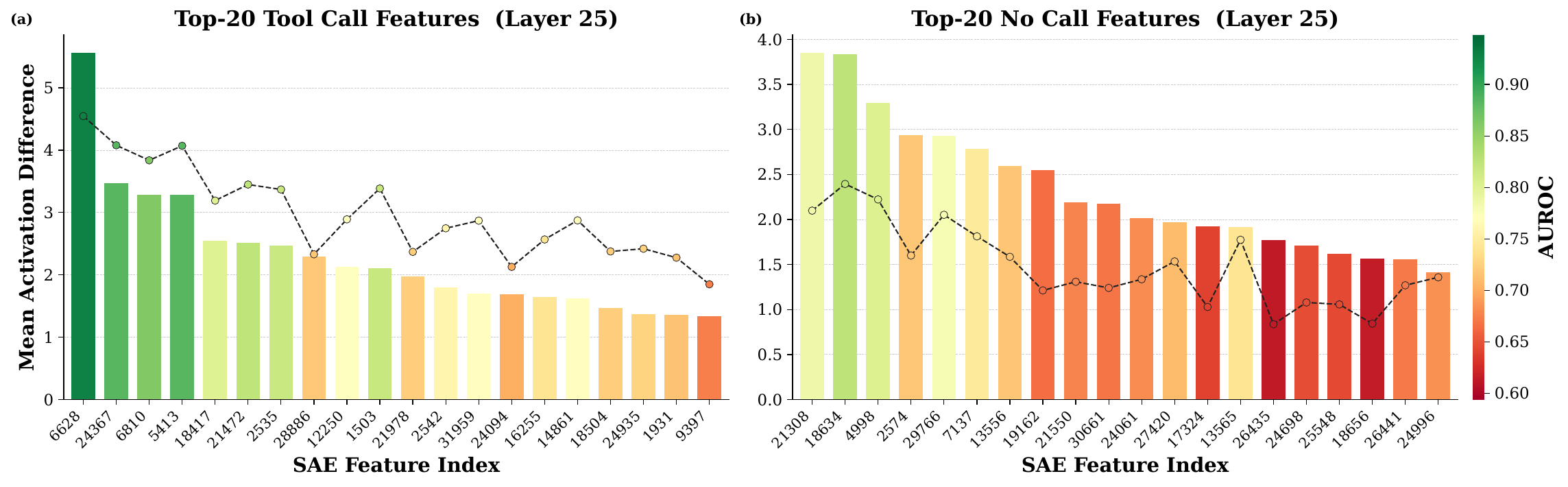}
  \caption{Top-ranked gating features discovered for Qwen3.5-9B. Bars show the mean activation difference between the target and contrast behavior-labeled sets, and the dashed line shows directional AUROC. Left: features associated with observed \textsc{call} decisions. Right: features associated with observed \textsc{no\_call} decisions.}
  \label{fig:feature-discovery}
  \vspace{-14pt}
\end{figure}

\paragraph{Feature extraction and ranking.}
For each $x_i \in \mathcal{D}$, we extract the residual-stream activation $\mathbf{h}_i$ at the action-boundary position, \ie the position of the final prompt token before the first generated response token. 
We encode $\mathbf{h}_i$ with Eq.~\eqref{equ:sae_encoder} to obtain SAE activation $\mathbf{z}_i \in \mathbb{R}^M$.
We first identify \textsc{call} features by treating $\mathcal{D}^{+}$ as the target set and $\mathcal{D}^{-}$ as the reference set.
For each SAE feature $j$, we compute the mean activation gap and the directional AUROC:
\begin{align}
  \Delta\mathrm{CE}_{\mathcal{C}}(j)
  &\;=\;
  \mathbb{E}_{\mathcal{D}^{+}}\!\left[z_j\right]
  \;-\;
  \mathbb{E}_{\mathcal{D}^{-}}\!\left[z_j\right], \label{equ:delta_ce} \\
  \mathrm{AUROC}_{\mathcal{C}}(j)
  &\;=\;
  P\!\left(z_j(x^{+}) > z_j(x^{-})\right),
  \quad x^{+} \sim \mathcal{D}^{+},\; x^{-} \sim \mathcal{D}^{-}.
  \label{equ:auroc}
\end{align}
For a ranking cutoff $R$, we keep the intersection of the top-$R$ features under the two scores.
The intersection favors features that are both strongly separated in mean activation and consistently discriminative across examples.
The \textsc{call} pass yields $\mathcal{C}$.
We obtain the \textsc{no\_call} feature set $\mathcal{N}$ by the same procedure after swapping the target and reference sets, using $\mathcal{D}^{-}$ against $\mathcal{D}^{+}$.
Sweeping $R$ gives the accuracy-sparsity trade-off used in validation.

Figure~\ref{fig:feature-discovery} shows the resulting top-ranked features for Qwen3.5-9B.
Both panels exhibit features with large activation gaps and high directional AUROC, indicating that the discovered features are not artifacts of a small number of extreme activations.
Analogous results for the remaining target models are reported in Appendix~\ref{app:feature_discovery_all_models}.

\begin{figure}[h]
  \centering
  \includegraphics[width=\linewidth]{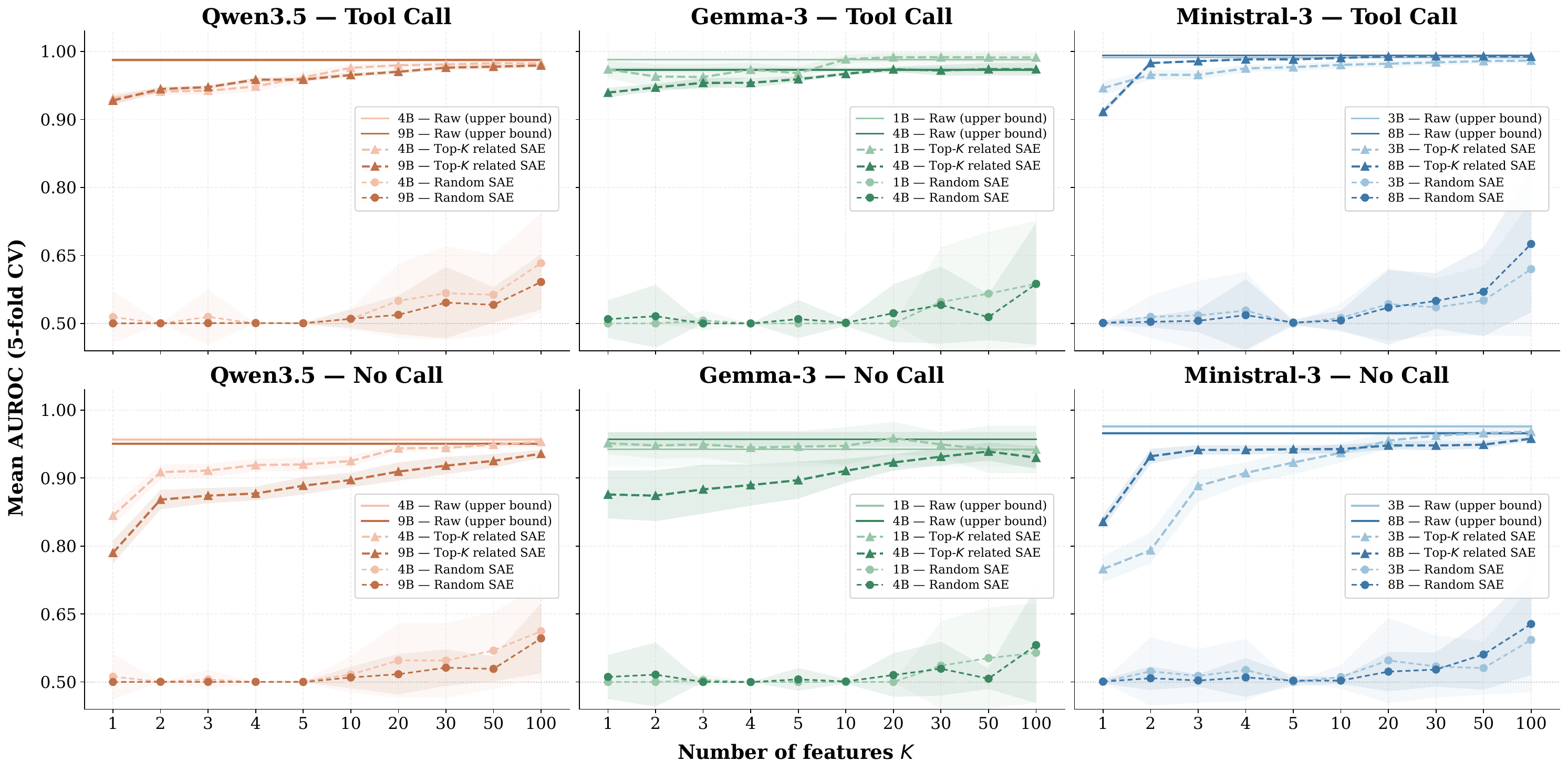}
  \caption{Discriminative validation across six target models. We train 5-fold cross-validated logistic probes with increasing numbers of selected SAE features and report mean AUROC. }
  \label{fig:discovery-probing}
  \vspace{-14pt}
\end{figure}
\subsection{Discriminative Validation}
\label{sec:validation}

\paragraph{Probe setup.}
We validate the discovered features by testing whether a small linear probe trained on them predicts the model's behavior-labeled gating decision.
For a selected feature set $\mathcal{S}$, we train a logistic regression on the sparse activations $\mathbf{z}_{\mathcal{S},\,i} = (z_j)_{j \in \mathcal{S}}$ at the action-boundary position:
\begin{equation}
  \hat{d}_i \;=\; \sigma\!\left(\mathbf{w}^{\top}\mathbf{z}_{\mathcal{S},\,i} + b\right),
  \label{equ:probe}
\end{equation}
where $\sigma$ is the sigmoid function.
Under 5-fold cross-validation, we compare three feature inputs: the top-ranked discovered SAE features, count-matched random SAE features, and the raw residual stream $\mathbf{h}_i$, where the last serves as an upper bound.
We sweep the number of selected features $|\mathcal{S}|$ and report mean AUROC across folds.

\textbf{Observation 1: A handful of discovered features predict the gating decision near the residual-stream upper bound.}
Figure~\ref{fig:discovery-probing} plots mean AUROC as $|\mathcal{S}|$ grows.
Across all three model families and both call and no-call directions, the discovered features reach the raw-residual upper bound with only a handful of SAE dimensions (typically $K \le 5$), while count-matched random features stay near chance.
This proximity to the upper bound indicates that $\mathcal{C}$ and $\mathcal{N}$ capture nearly all of the linearly recoverable gating signal in the residual stream, establishing them as a reliable feature basis.
Section~\ref{sec:diagnosis} uses them to test whether the over-calling bias reduces to activation levels alone.

\section{Diagnosing Intrinsic Decision Bias}
\label{sec:diagnosis}

This section diagnoses the mechanism of over-calling by asking whether the model's \textsc{call} preference is fully accounted for by \textsc{call}- and \textsc{no\_call}-feature activation levels, and by quantifying any activation-independent component of the decision mapping.

\subsection{Formalizing the Hypothesis}
\label{sec:hypothesis_formalization}

Section~\ref{sec:discovery} identifies two SAE feature sets, $\mathcal{C}$ and $\mathcal{N}$, that track the model's \textsc{call} and \textsc{no\_call} decisions.
Using these features, we ask whether over-calling is fully explained by activation levels, or whether the decision mapping itself carries an activation-independent \textsc{call} bias.
To make this question testable, we summarize the two feature groups by a signed activation margin and examine the model's decision as a function of that margin.

For each example $i$, let $z_{j,i}$ denote the activation of SAE feature $j$ at the action-boundary position.
Because SAE decoder columns are unit-norm by construction, feature mean activation can be compared directly in the SAE coordinate space.
We define the signed activation margin as the difference between \textsc{call}- and \textsc{no\_call}-feature mean activation:
\begin{equation}
  m_i = a_{\mathcal{C},i} - a_{\mathcal{N},i} = \frac{1}{|\mathcal{C}|}\sum_{j \in \mathcal{C}} z_{j,i} - \frac{1}{|\mathcal{N}|}\sum_{j \in \mathcal{N}} z_{j,i}.
\label{equ:activation_margin}
\end{equation}
where positive $m_i$ indicates stronger \textsc{call}-feature mean activation and negative $m_i$ the reverse.
Let $\hat{d}_i=1$ denote that the model's response is judged as \textsc{call}, and $\hat{d}_i=0$ otherwise.
The margin gives a direct diagnostic axis on which the two accounts make opposite predictions, formalized below.

\begin{tcolorbox}[colback=gray!4,colframe=black!35,boxrule=0.4pt,arc=1pt,left=5pt,right=5pt,top=4pt,bottom=4pt]
\textbf{Activation-only account ($H_{\mathrm{act}}$).}
The gating decision depends on the activation margin alone:
\begin{equation}
  \Pr(\hat{d}_i = 1 \mid m_i) = \sigma(\beta m_i),
  \qquad \beta > 0.
\label{equ:activation_only_decision}
\end{equation}
\textbf{Prediction.}
At activation parity, $\Pr(\hat{d}_i=1 \mid m_i=0)=1/2$.
The neutral decision boundary lies at $m_i=0$.
\end{tcolorbox}

\begin{tcolorbox}[colback=gray!4,colframe=black!35,boxrule=0.4pt,arc=1pt,left=5pt,right=5pt,top=4pt,bottom=4pt]
\textbf{Intrinsic Bias Hypothesis ($H_{\mathrm{IBH}}$).}
The decision mapping contains an activation-independent \textsc{call} offset:
\begin{equation}
  \Pr(\hat{d}_i = 1 \mid m_i) = \sigma(\beta m_i + \beta_0),
  \qquad \beta > 0,\quad \beta_0 > 0.
\label{equ:intrinsic_bias_decision}
\end{equation}
\textbf{Prediction.}
At activation parity, $\Pr(\hat{d}_i=1 \mid m_i=0)=\sigma(\beta_0)>1/2$.
The neutral boundary shifts to $m^{\star}=-\beta_0/\beta<0$.
\end{tcolorbox}

The two accounts therefore disagree on where the model becomes decision-neutral: at activation parity under $H_{\mathrm{act}}$, only after \textsc{no\_call} evidence exceeds \textsc{call} evidence under $H_{\mathrm{IBH}}$.
We first examine the activation geometry of \textsc{call}- and \textsc{no\_call}-feature evidence from both response- and input-conditioned views, then estimate the offset term $\beta_0$ directly.

\subsection{Evidence for Intrinsic Bias}

\begin{figure}[t]
  \centering
  \includegraphics[width=\linewidth]{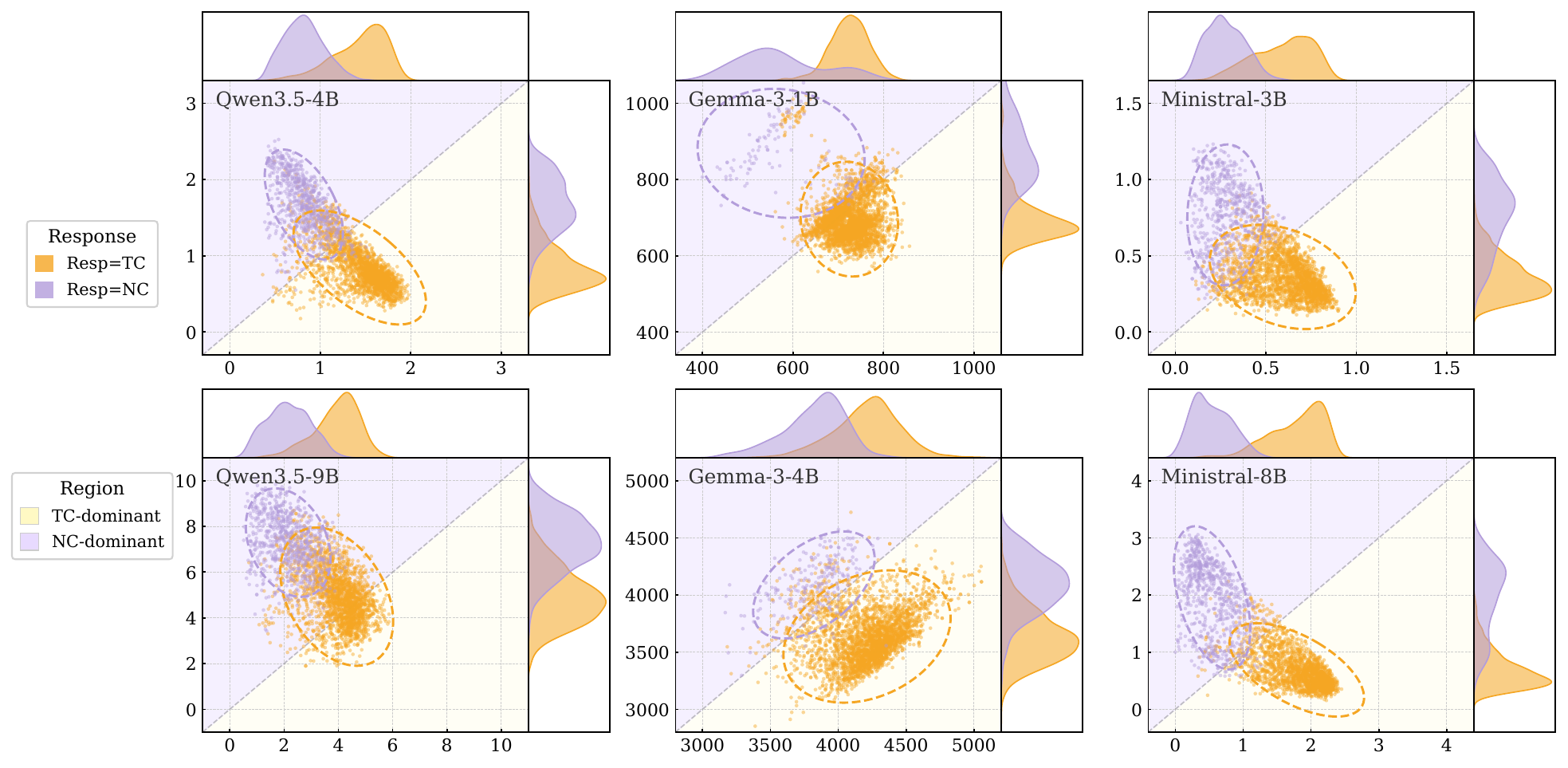}
  \caption{Response-conditioned activation geometry. Examples are grouped by the model's emitted decision: \textsc{no\_call} responses (\textcolor[HTML]{9F88BE}{purple}) stay in the \textsc{no\_call}-dominant half-plane, while \textsc{call} responses (\textcolor[HTML]{E89B45}{orange}) extend across the parity diagonal into the same region. The response boundary therefore sits on the \textsc{no\_call} side of the diagonal rather than at activation parity.}
  \label{fig:diagnosis-pred-tc-rfi}
  \vspace{-9pt}
\end{figure}

\begin{figure}[t]
  \centering
  \includegraphics[width=\linewidth]{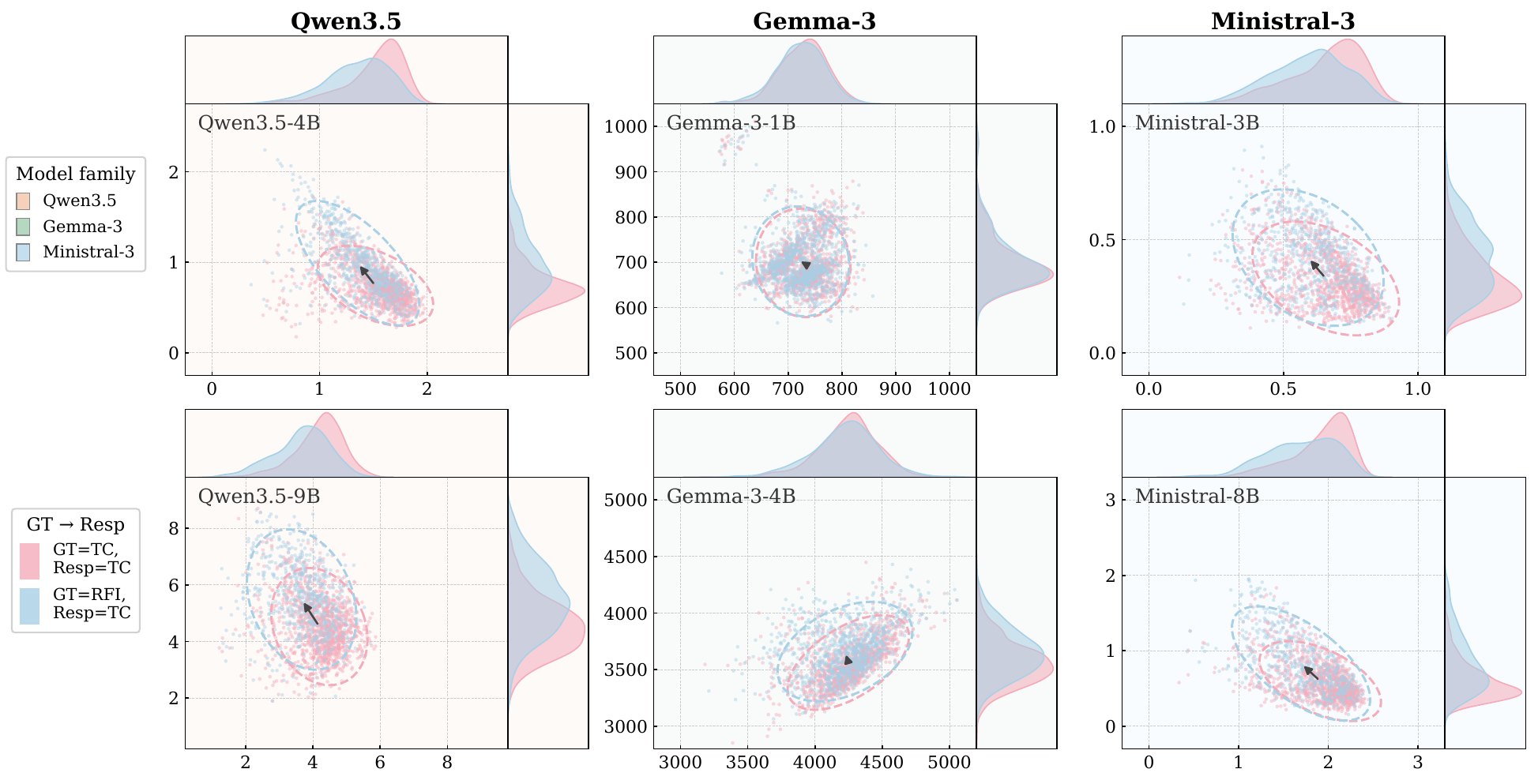}
  \caption{Input-conditioned activation geometry, restricted to examples for which the model emits a \textsc{call}. True calls (\textcolor[HTML]{E89BB5}{pink}) lie in the \textsc{call}-dominant region, while false calls (\textcolor[HTML]{6FA8DC}{blue}) overlap the same region but sit consistently closer to the parity diagonal. True and false calls therefore remain separable along the $\mathcal{C}$ and $\mathcal{N}$ axes within a single emitted decision.}
  \label{fig:diagnosis-pred-tc}
  \vspace{-14pt}
\end{figure}

To test these predictions, we examine each example's feature evidence in the plane $(a_{\mathcal{C}}, a_{\mathcal{N}})$, where the diagonal marks $m=0$.
We probe this geometry from two complementary views.
A response-conditioned view asks whether the parity line separates emitted \textsc{call} from \textsc{no\_call} responses.
An input-conditioned view then asks whether, among emitted \textsc{call} responses, true and false calls share the same feature geometry.

\textbf{Observation 2: The response boundary is shifted into the \textsc{no\_call}-dominant half-plane.}
Across all six target models (Figure~\ref{fig:diagnosis-pred-tc-rfi}), \textsc{no\_call} responses concentrate well inside the \textsc{no\_call}-dominant half-plane, with the model emitting \textsc{no\_call} only when \textsc{no\_call}-feature mean activation clearly exceeds \textsc{call}-feature mean activation.
\textsc{call} responses, by contrast, extend across the parity line into regions where \textsc{no\_call} features are comparable to or stronger than \textsc{call} features.
This shift is systematic: the model carries a \textsc{call} preference that persists even when feature evidence already favors \textsc{no\_call}.
This is the directional signature of $H_{\mathrm{IBH}}$.

\textbf{Observation 3: Among emitted calls, false calls lie closer to the \textsc{no\_call} side than true calls.}
Conditioning on emitted \textsc{call} responses (Figure~\ref{fig:diagnosis-pred-tc}), true and false calls occupy distinct activation distributions: across all six models, the false-call mass sits consistently closer to the parity diagonal than the true-call mass.
The same intrinsic \textsc{call} preference is therefore still acting inside the \textsc{call} subset, pushing examples with weaker \textsc{call} evidence over the response boundary.
Crucially, true and false calls remain separable along the $\mathcal{C}$ and $\mathcal{N}$ axes, giving Section~\ref{sec:steering} a directional lever: a counter-bias along the SAE decoder directions of $\mathcal{C}$ and $\mathcal{N}$ can push the shifted false-call mass back across the boundary while leaving the more deeply $\mathcal{C}$-dominant true calls in place.

Together, Observations 2 and 3 establish the qualitative shape of $H_{\mathrm{IBH}}$: the bias surfaces both in where the response boundary sits and in which emitted calls turn out wrong.
We now estimate $\beta_0$ to make this offset quantitative.

\subsection{Quantifying the Bias Offset}

\begin{figure}[h]
  \centering
  \includegraphics[width=\linewidth]{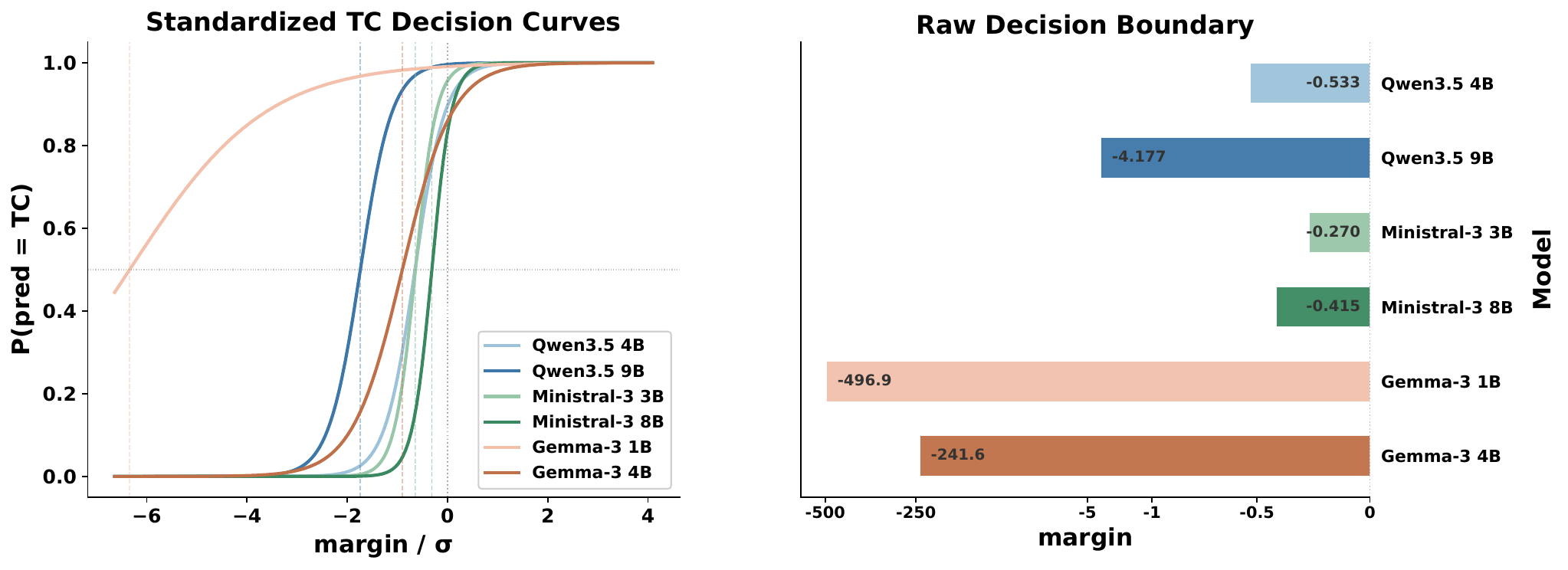}
  \caption{Logistic estimates of intrinsic \textsc{call} bias. \textbf{Left}: fitted decision curves with margin standardized within model, so the six curves can be visually compared. \textbf{Right}: raw neutral boundaries $\hat{m}^{\star}$. All boundaries are negative, meaning the model becomes decision-neutral only when \textsc{no\_call}-feature mean activation exceeds \textsc{call}-feature mean activation.}
  \label{fig:diagnosis-bias-summary}
  \vspace{-6pt}
\end{figure}

For each model, we fit Eq.~\ref{equ:intrinsic_bias_decision} on cached SAE activations and observed model decisions, yielding empirical estimates $(\hat{\beta}, \hat{\beta}_0)$.
The intercept $\hat{\beta}_0$ sets the \textsc{call} probability at activation parity, and the neutral boundary $\hat{m}^{\star}=-\hat{\beta}_0/\hat{\beta}$ marks the margin at which $\Pr(\hat{d}_i=1)=1/2$.
The left panel of Figure~\ref{fig:diagnosis-bias-summary} expresses margin in within-model standard-deviation units, aligning the six fitted curves: each crosses $0.5$ at a negative margin and sits above $0.5$ at activation parity.
The right panel reports the raw $\hat{m}^{\star}$, which are negative for all six models and span several orders of magnitude across families.
Across all six models, $\hat{m}^{\star}<0$ supports $H_{\mathrm{IBH}}$: parity between \textsc{call}- and \textsc{no\_call}-feature mean activations does not bring the model to decision-neutrality.
The much larger raw magnitudes for Gemma likely reflect the larger numerical scale of its activations, so $\hat{m}^{\star}$ should be read as a within-model bias measure rather than a cross-model strength.

\section{Causal Steering Experiments}
\label{sec:steering}

\newcommand{\dup}[2]{$#1_{\textcolor{myorange}{\scriptscriptstyle +#2}}$}
\newcommand{\ddn}[2]{$#1_{\textcolor{black!55}{\scriptscriptstyle -#2}}$}
\newcommand{\dno}[1]{\textit{#1}\hphantom{${}_{\scriptscriptstyle -99.99}$}}

If the offset $\beta_0$ in Section~\ref{sec:diagnosis} is part of the mechanism that produces over-calling, counteracting it along the same SAE feature directions should rebalance \textsc{call} and \textsc{no\_call} decisions and improve overall accuracy.
We instantiate this causal test with Adaptive Margin-Calibrated Steering (\textbf{AMCS}), a closed-form activation steering method derived from the fitted decision-margin model.

\subsection{Adaptive Margin-Calibrated Steering}
\label{sec:amcs}

\paragraph{Closed-form calibration.}
AMCS reuses the signed activation margin in Eq.~\eqref{equ:activation_margin}.
For a steering budget of $r$ features per side, let $\mathcal{C}_r \subseteq \mathcal{C}$ and $\mathcal{N}_r \subseteq \mathcal{N}$ denote the top-ranked \textsc{call} and \textsc{no\_call} features from Section~\ref{sec:discovery_pipeline}.
On cached calibration activations and observed model decisions, we refit the diagnostic margin model on this restricted basis using the same margin definition as Section~\ref{sec:hypothesis_formalization}:
\begin{equation}
  \Pr(\hat{d}=1 \mid m_r) = \sigma(\beta_r m_r + \beta_{0,r}),
  \label{equ:amcs_logistic}
\end{equation}
where $\hat{d}=1$ denotes an observed \textsc{call} response, $m_r$ is the signed margin recomputed on $\mathcal{C}_r \cup \mathcal{N}_r$, and $\beta_r,\beta_{0,r}$ are the fitted slope and \textsc{call} offset under budget $r$.
To remove this offset, AMCS shifts every margin by the same amount $\delta_r$ so that the calibrated logit matches the unbiased one, $\beta_r(m+\delta_r)+\beta_{0,r}=\beta_r m$ for all $m$. This yields the closed-form:
\begin{equation}
  \delta_r = -\frac{\beta_{0,r}}{\beta_r},
  \label{equ:amcs_delta}
\end{equation}
so the steering strength is fixed by the diagnosed bias rather than tuned on a validation set.

\paragraph{Steering vector and intervention.}
Let $\mathbf{d}_j$ be the SAE decoder column for feature $j$.
To allocate the shift across selected features, we measure each feature's activation gap between \textsc{call}-decision and \textsc{no\_call}-decision responses on the calibration set, $\Delta^{\mathcal{C}}_j$ for $j\in\mathcal{C}_r$ and $\Delta^{\mathcal{N}}_j$ for $j\in\mathcal{N}_r$, and normalize absolute gaps within each side:
\begin{equation}
  \omega^{\mathcal{C}}_j = \frac{|\Delta^{\mathcal{C}}_j|}{\sum_{k \in \mathcal{C}_r}|\Delta^{\mathcal{C}}_k|}, \qquad \omega^{\mathcal{N}}_j = \frac{|\Delta^{\mathcal{N}}_j|}{\sum_{k \in \mathcal{N}_r}|\Delta^{\mathcal{N}}_k|}.
  \label{equ:amcs_weights}
\end{equation}
These weights only set allocation, with the magnitude fixed by $\delta_r$. The steering vector is then:
\begin{equation}
  \mathbf{v}_r = \underbrace{\alpha r \delta_r \sum_{j \in \mathcal{C}_r}\omega^{\mathcal{C}}_j \mathbf{d}_j}_{\textsc{call}~suppression~(\delta_r<0)} + \underbrace{(1-\alpha) r (-\delta_r) \sum_{j \in \mathcal{N}_r}\omega^{\mathcal{N}}_j \mathbf{d}_j}_{\textsc{no\_call}~enhancement~(-\delta_r>0)},
  \label{equ:amcs_vector}
\end{equation}
where $\alpha \in [0,1]$ trades off the correction between \textsc{call}-side and \textsc{no\_call}-side decoder directions, and the factor $r$ cancels the per-side averaging in $m_r$ so that $\mathbf{v}_r$ targets a total margin shift of $\delta_r$ under a local linear SAE approximation.
At inference, we add $\mathbf{v}_r$ at the SAE hook layer throughout autoregressive generation, $\mathbf{H}_{\ell} \leftarrow \mathbf{H}_{\ell}+\mathbf{v}_r$, costing one broadcast addition per token. Implementation details are in Appendix~\ref{app:amcs_details}.

\subsection{Causal Validation Across Models}
\label{sec:steering_validation}

\paragraph{Setup.} We evaluate AMCS on When2Call across six instruction-tuned models from three families: Qwen3.5-(4B, 9B), Gemma-3-it-(1B, 4B), and Ministral-3-Instruct-(3B, 8B), reusing the SAE and diagnosis layer of Section~\ref{sec:diagnosis}. 
AMCS fits $\beta_r$ and $\beta_{0,r}$ on a calibration split and is evaluated on a held-out test split, with the balance set to $\alpha=0.8$.
We scan the steering budget $r \in \{5, 10, 15, 20, 25, 30\}$ and report the mean across $r$.
We compare against three reference interventions, Prompt, Suppress, and Promote, defined in Table~\ref{tab:steering_validation}.
The latter two each cover only one half of the AMCS shift, isolating the effect of the closed-form two-sided calibration.

\paragraph{Results.} Each of the three reference interventions fails in a characteristic way (Table~\ref{tab:steering_validation}).
Prompt acts at the surface and overshoots: on Ministral-3-8B it raises no-call accuracy by 45 points but cuts tool-call accuracy by 56 points, dropping Overall below \textit{Init}.
Suppress and Promote act in the right space but only on one side of the margin: on Qwen3.5-4B, each moves no-call accuracy by under 10 points and Overall by under 3.2 points.
AMCS, by combining both sides with magnitude set by $\beta_0$, keeps tool-call accuracy within 5 points of \textit{Init} on five of six models\footnote{On the two Gemma-3 models, true and false calls show little separation along the $\mathcal{C}$ and $\mathcal{N}$ axes (Figure~\ref{fig:diagnosis-pred-tc}), so the SAE feature basis offers little leverage for any margin-based intervention. Gemma-3-4B is a null case, with Overall nearly unchanged, and Gemma-3-1B is a failure case, with Overall falling by 1.64 points. AMCS is therefore a model-conditional intervention.} while raising no-call accuracy by 4 to 17 points, giving the best Overall on Qwen3.5-4B, Qwen3.5-9B, and Ministral-3-3B, and landing within 0.6 points of the best on Ministral-3-8B.
These gains come from cancelling the diagnosed offset $\beta_0$ along the same SAE directions, the causal counterpart of IBH.

\begin{table}[t]
\centering
\caption{Causal validation of AMCS on When2Call (\%). TC Acc and NC Acc denote tool-call and no-call accuracy. \textit{Init} is the unmodified model. Prompt appends ``When the user's request lacks necessary details, ask before taking action.'' to the user prompt. Suppress scales the top-ranked \textsc{call} feature activations by $0.5$. Promote scales the top-ranked \textsc{no\_call} feature activations by $1.5$. Subscripts give the absolute change versus \textit{Init} in percentage points (\textcolor{myorange}{positive} / \textcolor{black!55}{negative}).}
\label{tab:steering_validation}
\small
\setlength{\tabcolsep}{4pt}
\resizebox{\textwidth}{!}{%
\begin{tabular}{llcccccc}
\toprule
Metric & Method & Qwen3.5-4B & Qwen3.5-9B & Gemma3-1B & Gemma3-4B & Ministral3-3B & Ministral3-8B \\
\midrule
\multirow{5}{*}{\textbf{TC Acc}}
 & \textit{Init}    & \dno{95.98} & \dno{96.45} & \dno{98.61} & \dno{99.23} & \dno{91.66} & \dno{91.97} \\
 & Prompt                   & \ddn{68.11}{27.87} & \ddn{67.72}{28.73} & \ddn{94.67}{3.94} & \ddn{58.22}{41.01} & \ddn{61.62}{30.04} & \ddn{35.91}{56.06} \\
 & Suppress         & \ddn{93.48}{2.50} & \ddn{94.61}{1.84} & \ddn{93.84}{4.77} & \ddn{94.34}{4.89} & \ddn{86.75}{4.91} & \ddn{85.34}{6.63} \\
 & Promote          & \ddn{93.82}{2.16} & \dup{96.88}{0.43} & \ddn{98.38}{0.23} & \ddn{98.53}{0.70} & \ddn{90.07}{1.59} & \ddn{89.92}{2.05} \\
 & Ours                     & \ddn{91.82}{4.16} & \ddn{91.72}{4.73} & \ddn{89.77}{8.84} & \ddn{99.18}{0.05} & \ddn{86.66}{5.00} & \ddn{89.90}{2.07} \\
\midrule
\multirow{5}{*}{\textbf{NC Acc}}
 & \textit{Init}    & \dno{37.76} & \dno{37.57} & \dno{2.26} & \dno{5.56} & \dno{34.65} & \dno{41.71} \\
 & Prompt                   & \dup{77.31}{39.55} & \dup{79.38}{41.81} & \dup{15.91}{13.65} & \dup{59.23}{53.67} & \dup{71.56}{36.91} & \dup{87.10}{45.39} \\
 & Suppress         & \dup{44.65}{6.89} & \dup{44.43}{6.86} & \dup{4.35}{2.09} & \dup{7.65}{2.09} & \dup{45.75}{11.10} & \dup{52.75}{11.04} \\
 & Promote          & \dup{47.36}{9.60} & \ddn{34.27}{3.30} & \dup{2.97}{0.71} & \dup{7.64}{2.08} & \dup{41.26}{6.61} & \dup{46.91}{5.20} \\
 & Ours                     & \dup{51.07}{13.31} & \dup{54.58}{17.01} & \dup{9.40}{7.14} & \ddn{5.55}{0.01} & \dup{45.91}{11.26} & \dup{45.76}{4.05} \\
\midrule
\multirow{5}{*}{\textbf{Overall}}
 & \textit{Init}    & \dno{69.75} & \dno{69.88} & \dno{55.20} & \dno{57.11} & \dno{65.97} & \dno{69.33} \\
 & Prompt                   & \dup{72.25}{2.50} & \dup{72.97}{3.09} & \dup{\textbf{59.19}}{3.99} & \dup{\textbf{58.68}}{1.57} & \dup{66.10}{0.13} & \ddn{58.97}{10.36} \\
 & Suppress         & \dup{71.48}{1.73} & \dup{72.00}{2.12} & \ddn{53.52}{1.68} & \ddn{55.28}{1.83} & \dup{68.27}{2.30} & \dup{\textbf{70.66}}{1.33} \\
 & Promote          & \dup{72.89}{3.14} & \ddn{68.67}{1.21} & \dup{55.39}{0.19} & \dup{57.58}{0.47} & \dup{68.08}{2.11} & \dup{70.54}{1.21} \\
 & Ours                     & \dup{\textbf{73.47}}{3.72} & \dup{\textbf{74.98}}{5.10} & \ddn{53.56}{1.64} & \ddn{57.00}{0.11} & \dup{\textbf{68.30}}{2.33} & \dup{70.10}{0.77} \\
\bottomrule
\end{tabular}%
}
\vspace{-10pt}
\end{table}

\section{Transfer and Robustness}
\label{sec:transferability}

This section asks whether the offset and intervention transfer to another benchmark, whether AMCS preserves valid calls, and whether the IBH offset depends on the SAE and margin choices used above.
Detailed results are in Appendix~\ref{app:transfer_details}, and results on models above 10B, a pretrained checkpoint, and reasoning-enabled decoding are in Appendix~\ref{app:additional_settings}.

\subsection{Cross-Benchmark Transfer}
\label{sec:acebench}

We transfer the diagnosis and the intervention to a 100-example ACEBench~\cite{acebench} set, with the When2Call SAE, \textsc{call}/\textsc{no\_call} features, and steering directions held fixed.
The split, the no-call scoring rule, margin normalization, and other details are in Appendix~\ref{app:acebench}.

The frozen basis reproduces IBH on the four models whose \textsc{call}/\textsc{no\_call} boundary is identifiable.
For Qwen3.5-4B, Qwen3.5-9B, Ministral-3-3B, and Ministral-3-8B, the 95\% bootstrap interval for $\Pr(\hat{d}=1\mid m=0)$ lies above $0.5$, and the interval for $m^{\star}$ lies below zero.
Gemma-3-1B and Gemma-3-4B yield 99/0 and 98/2 \textsc{call}/\textsc{no\_call} decisions, so the parity regression is not estimable.

Steering is evaluated on these four models.
Gemma is excluded because its \textsc{call}/\textsc{no\_call} separation is already weak on When2Call (Section~\ref{sec:steering_validation}) and its ACEBench decisions collapse to tool calls.
Contrastive Activation Addition (\textbf{CAA})~\cite{caa} uses the same hook layer, with its coefficient selected separately per model on When2Call.
AMCS uses one frozen configuration, $r=25$ and $\alpha=0.8$, for all four models.
Neither method uses ACEBench labels to choose its direction or its strength.

\begin{table}[ht]
  \centering
  \caption{Frozen transfer to ACEBench (\%). Each cell is overall / call / no-call accuracy. Baselines follow Table~\ref{tab:steering_validation}. CAA uses a per-model coefficient selected on When2Call, and AMCS uses $r=25$, $\alpha=0.8$ for all four models. \textbf{Bold} and \underline{underline} mark the best and second best macro averages.}
  \label{tab:acebench}
  \small
  \setlength{\tabcolsep}{4pt}
  \begin{tabular}{lccccc}
    \toprule
    Method & Qwen3.5-4B & Qwen3.5-9B & Ministral-3-3B & Ministral-3-8B & Macro \\
    \midrule
    Init     & 76 / 98 / 54 & 70 / 100 / 40 & 81 / 94 / 68 & 85 / 98 / 72 & 78.00 / \textbf{97.50} / 58.50 \\
    Prompt   & 82 / 72 / 92 & 91 / 92 / 90 & 86 / 76 / 96 & 69 / 42 / 96 & 82.00 / 70.50 / \textbf{93.50} \\
    Suppress & 79 / 94 / 64 & 76 / 100 / 52 & 84 / 92 / 76 & 85 / 90 / 80 & 81.00 / 94.00 / 68.00 \\
    Promote  & 85 / 98 / 72 & 75 / 100 / 50 & 85 / 94 / 76 & 87 / 96 / 78 & \underline{83.00} / \underline{97.00} / 69.00 \\
    CAA      & 84 / 98 / 70 & 77 / 100 / 54 & 83 / 92 / 74 & 87 / 96 / 78 & 82.75 / 96.50 / 69.00 \\
    AMCS     & 84 / 98 / 70 & 80 / 98 / 62 & 86 / 92 / 80 & 85 / 96 / 74 & \textbf{83.75} / 96.00 / \underline{71.50} \\
    \bottomrule
  \end{tabular}
\end{table}

Relative to Init, frozen AMCS raises macro-average overall accuracy from 78.00\% to 83.75\% and no-call accuracy from 58.50\% to 71.50\%, while call accuracy changes from 97.50\% to 96.00\%.
Prompt reaches the highest no-call accuracy but lowers call accuracy to 70.50\%.
CAA, with a coefficient tuned per model, reaches 82.75 / 96.50 / 69.00.
AMCS is not better on every model: it ties CAA on Qwen3.5-4B, leads by 3 points on Qwen3.5-9B and Ministral-3-3B, and trails by 2 points on Ministral-3-8B.
With 100 examples per model these margins are small, but a configuration fixed on When2Call still raises overall and no-call accuracy with call accuracy close to Init.

\subsection{Preservation of Valid Calls}
\label{sec:preserved_calls}

Call accuracy only records whether a tool is called, so we also check whether AMCS keeps valid calls correct.
On all 1,295 ground-truth tool-call examples in When2Call, we pair each unsteered response with its AMCS response and score function names and arguments with a schema-conditioned judge (Appendix~\ref{app:preserved_calls}).
Across the five models other than Gemma-3-1B, AMCS preserves 95.73\% of the unsteered tool calls, and 97.3\% of the flips to \textsc{no\_call} fall in the lower-margin half.
On the calls preserved by both responses, joint function-and-argument accuracy changes by $-0.27$ points on average.
Gemma-3-1B is a failure case: even a reduced Top-6 intervention flips 24.18\% of its tool calls and lowers joint accuracy by 20.87 points.

\subsection{Robustness to SAE and Margin Choices}
\label{sec:robustness}

A setting supports IBH when $\beta>0$, $\Pr(\hat{d}=1\mid m=0)>0.5$, and $m^{\star}<0$.
For each of the six models we retrain five additional SAEs, changing one factor at a time (two hook layers, $K/2$, $2K$, and a second random seed), and rediscover Top-25 features in each new basis.
All six settings for Qwen3.5-4B support IBH (Table~\ref{tab:sae_ablation_qwen}), and 34 of the 36 settings across the six models do.
The two exceptions, Ministral-3-3B at Layer 24 and Ministral-3-8B at $K/2$, keep a positive slope but yield a slightly positive $m^{\star}$.
On the original SAE, IBH also holds in 35 of 36 feature-budget settings and in all 18 margin-construction settings.
We therefore claim robustness to the tested SAE and margin choices (Appendix~\ref{app:robustness}), not invariance to every configuration.

\begin{table}[t]
  \centering
  \caption{SAE ablation for Qwen3.5-4B. Each row refits the diagnostic logistic model after rediscovering Top-25 features in that SAE. Original is the setting of Section~\ref{sec:diagnosis}. Raw $m^{\star}$ is not comparable across rows.}
  \label{tab:sae_ablation_qwen}
  \small
  \setlength{\tabcolsep}{5pt}
  \begin{tabular}{lrrrccc}
    \toprule
    Setting & Layer & $K$ & Seed & $\beta$ & $\Pr(\hat{d}=1\mid m=0)$ & $m^{\star}$ \\
    \midrule
    Original & 25 & 80 & 42 & 4.006 & 0.894 & $-0.533$ \\
    Layer 22 & 22 & 80 & 42 & 4.652 & 0.742 & $-0.227$ \\
    Layer 28 & 28 & 80 & 42 & 3.574 & 0.903 & $-0.626$ \\
    $K/2$ & 25 & 40 & 42 & 3.209 & 0.903 & $-0.695$ \\
    $2K$ & 25 & 160 & 42 & 4.203 & 0.935 & $-0.634$ \\
    Seed 123 & 25 & 80 & 123 & 3.901 & 0.939 & $-0.703$ \\
    \bottomrule
  \end{tabular}
\end{table}

\section{Conclusion}
\label{sec:conclusion}

Tool-using LLM agents must decide not only how to call tools, but also when not to call.
This paper studied a systematic failure in that decision: models achieve high call accuracy while remaining much less reliable on no-call cases, producing over-calling.
Using SAE feature bases for the \textsc{call}/\textsc{no\_call} gating decision, we showed that this asymmetry is not fully explained by feature activation levels.
We formalized this as IBH: an activation-independent call offset that shifts the neutral boundary toward \textsc{call}, so even at activation parity the decision remains biased toward calling.

We then turned this diagnosis into AMCS, a closed-form steering method that counteracts the fitted offset along SAE decoder directions.
The same offset remains under the tested SAE settings, and the intervention transfers to another benchmark with its directions held fixed.
This unifies behavioral miscalibration, feature-level mechanism, and causal intervention.
Over-calling is therefore not only an empirical artifact of tool-use benchmarks.
It is a mechanistic object that can be measured, modeled, and causally adjusted.
Limitations and further discussion are in Appendix~\ref{sec:discussions}.

\clearpage
\bibliography{neurips_2026}
\bibliographystyle{unsrt}
\newpage
\appendix

\section{Limitations and Discussions}
\label{sec:discussions}

\paragraph{Limitations.}
Appendix~\ref{app:base_checkpoint} compares Qwen3.5-4B with its official pretrained checkpoint.
The offset is already detectable before post-training, but the comparison does not identify which training stage produces it.
AMCS remains an inference-time correction, not a training-side fix.
The intervention relies on a local linear approximation that translates decoder-direction shifts into margin shifts.
The experiments in this paper cover only single-turn decisions, and we leave multi-turn interactions to future work.

\paragraph{Toward controllable calling effort.}
Modern LLM interfaces expose knobs for reasoning effort or compute budget, but tool use lacks a comparable control over how readily a model invokes external actions rather than verifying intent with the user.
The diagnosed call offset behaves like such a knob: shifting it tunes the balance between eager calling and seeking user verification, without altering the underlying tool-use capability.
AMCS is therefore not only a correction for over-calling, but a first step toward a ``calling effort'' interface where users or systems can set a task-specific point on this call-versus-verify axis.

\paragraph{Toward agent interpretability.}
Interpreting agents requires more than explaining isolated outputs, since real agents act over multiple turns, condition on tool feedback, and update plans dynamically.
This paper isolates one consequential unit of that broader problem: the call-versus-verify decision.
We connect it to feature-level geometry and a causal intervention, so the mechanism can be measured and adjusted.
Extending this analysis to long-horizon agents will require tracking such mechanisms across time, memory, and tool outputs.

\section{LLM as Judge Prompt}
\label{app:judge_prompt}

\begin{promptbox}
\raggedright
You are an expert at classifying responses from AI models.

Your task is to classify AI model's response into one of the following
four categories:
\begin{enumerate}[leftmargin=*, label=(\arabic*)]
\item \texttt{direct\_answer}: The AI model responded to the User's
      questions based on its existing knowledge, without requesting
      any additional information or using external tools.
\item \texttt{tool\_call}: The AI model decided to use a tool from
      the provided ones to help answer the question.
\item \texttt{request\_for\_info}: The AI model requested
      additional information from the User.
\item \texttt{cannot\_answer}: The AI model refused to answer the
      User's questions by acknowledging the lack of required
      capabilities.
\end{enumerate}

\textit{You should not judge whether the AI model's response is
accurate or not. Only provide the classification into one of these
four categories.}

\medskip
\texttt{<AVAILABLE\_TOOLS>} \{tools\} \texttt{</AVAILABLE\_TOOLS>}\\
\texttt{<USER\_QUESTION>} \{question\} \texttt{</USER\_QUESTION>}\\
\texttt{<AI\_MODEL\_RESPONSE>} \{response\} \texttt{</AI\_MODEL\_RESPONSE>}

\medskip
Please provide the classification in the following JSON format:\\
\texttt{\{"classification": "<direct\_answer | tool\_call |
request\_for\_info | cannot\_answer>"\}}

Respond only in the prescribed JSON format.
\end{promptbox}

\section{Behavior-Labeled Discovery Set}
\label{app:behavior_counts}

\begin{table}[h]
  \centering
  \small
  \caption{Per-model response counts across the four When2Call categories assigned by the LLM judge. We form $\mathcal{D}^{+}$ from \texttt{tool\_call} responses and $\mathcal{D}^{-}$ from \texttt{request\_for\_info} responses, and exclude the remaining two categories from the discovery set.}
  \label{tab:behavior_counts}
  \begin{tabular}{lrrrr}
    \toprule
    Model & \texttt{tool\_call} & \texttt{request\_for\_info} & \texttt{direct\_answer} & \texttt{cannot\_answer}  \\
    \midrule
    Qwen3.5-4B               & 2163 & 751 & 272 & 466 \\
    Qwen3.5-9B               & 2151 & 758 & 257 & 486 \\
    Gemma-3-1b-it            & 3318 & 104 & 47  & 173 \\
    Gemma-3-4b-it            & 3037 & 223 & 207 & 185 \\
    Ministral-3-3B-Instruct  & 2408 & 603 & 363 & 278 \\
    Ministral-3-8B-Instruct  & 1993 & 731 & 382 & 546 \\
    \bottomrule
  \end{tabular}
\end{table}

We build the discovery set from the full When2Call evaluation split, running each target model on all contexts regardless of the original category.
The LLM judge classifies each response into one of the four When2Call response types: \texttt{tool\_call}, \texttt{request\_for\_info}, \texttt{direct\_answer}, and \texttt{cannot\_answer}.
We then form $\mathcal{D}^{+}$ from \texttt{tool\_call} responses and $\mathcal{D}^{-}$ from \texttt{request\_for\_info} responses. Responses in the other two categories describe plain answering and tool unavailability rather than the call/no-call gating decision and are excluded from the discovery set.

Table~\ref{tab:behavior_counts} reports per-model counts of judged responses across the four categories.
The first two columns correspond to $|\mathcal{D}^{+}|$ and $|\mathcal{D}^{-}|$, whose cross-model variation reflects each model's response distribution.

\FloatBarrier
\section{SAE Training Details}
\label{app:sae_training}

\begin{table}[h]
  \centering
  \small
  \caption{SAE training configuration for each target model. $d$ is the text-backbone residual width, $L$ is the number of transformer blocks, $\ell$ is the zero-indexed hook block, $M=8d$ is the SAE dictionary size, and $K=\lfloor d/32 \rfloor$ is the TopK sparsity.}
  \label{tab:sae_training_config}
  \begin{tabular}{lrrrrr}
    \toprule
    Model & $d$ & $L$ & $\ell$ & $M$ & $K$ \\
    \midrule
    Qwen3.5-4B & 2560 & 32 & 25 & 20480 & 80 \\
    Qwen3.5-9B & 4096 & 32 & 25 & 32768 & 128 \\
    Gemma-3-1b-it & 1152 & 26 & 17 & 9216 & 36 \\
    Gemma-3-4b-it & 2560 & 34 & 29 & 20480 & 80 \\
    Ministral-3-3B-Instruct & 3072 & 26 & 21 & 24576 & 96 \\
    Ministral-3-8B-Instruct & 4096 & 34 & 31 & 32768 & 128 \\
    \bottomrule
  \end{tabular}
  \vspace{-6pt}
\end{table}

We train SAEs on the text-backbone residual stream of each target model.
Following the $M=8d$ and $K=\lfloor d/32 \rfloor$ configuration introduced in \S\ref{sec:sae_training}, we instantiate per-model dimensions as listed in Table~\ref{tab:sae_training_config}, where the hook block $\ell$ is chosen from the middle-to-late transformer blocks of each model under zero-based indexing.
Both training stages use AdamW \cite{adamw} with a learning rate of $5 \times 10^{-4}$, $\beta=(0.9, 0.999)$, a batch size of $16{,}384$ tokens, and a warmup--stable--decay schedule ($10\%$--$80\%$--$10\%$).

\begin{figure}[h]
  \centering
  \includegraphics[width=0.7\linewidth]{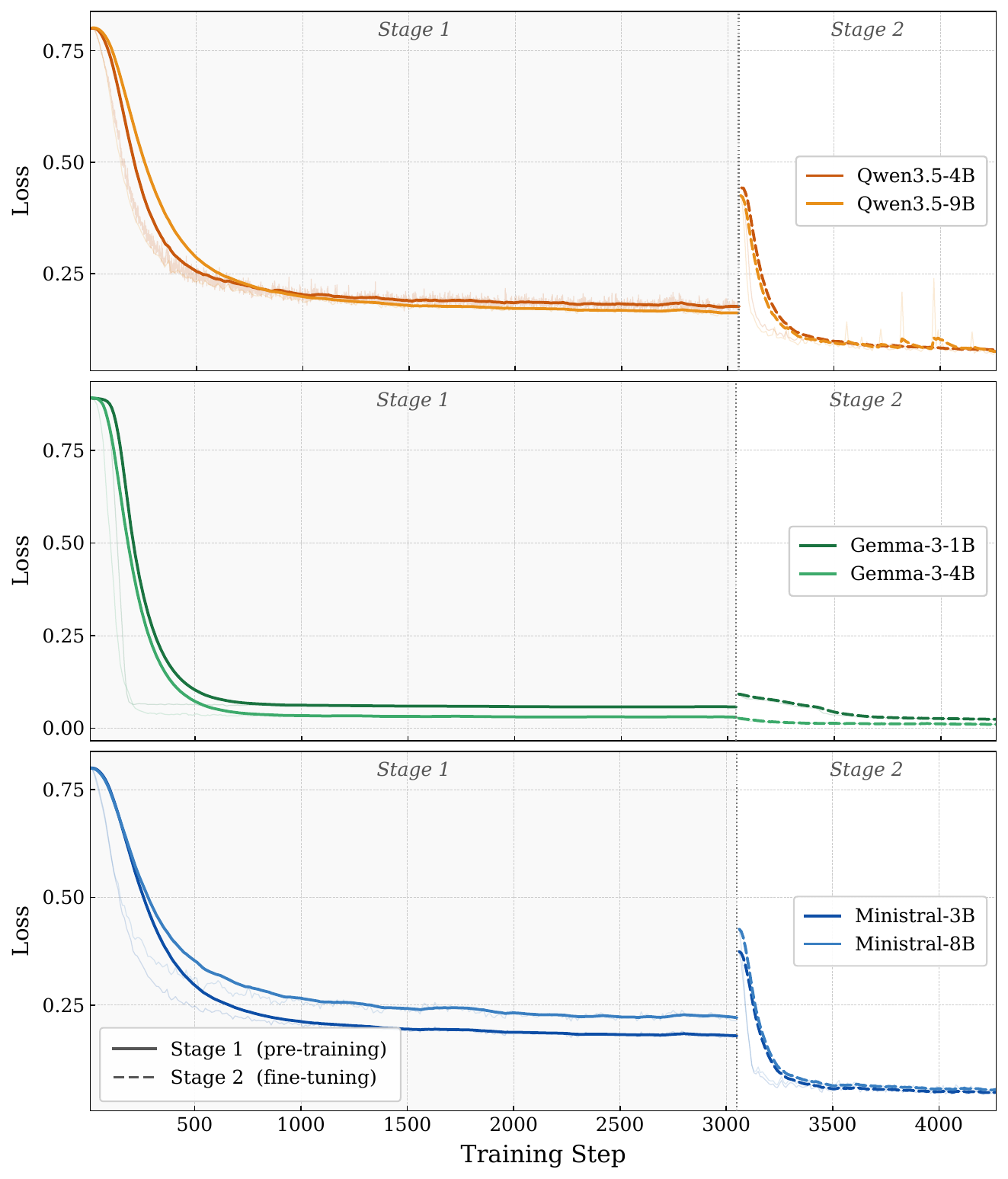}
  \caption{SAE reconstruction loss across the two-stage training curriculum, grouped by model family. Solid traces show Stage 1 (broad-corpus pre-training on OpenWebText2) and dashed traces show Stage 2 (When2Call adaptation), separated by the vertical divider. Light curves are raw per-step losses and bold curves are running averages. Within each panel, the two color shades distinguish the two model scales.}
  \label{fig:sae_training_diagnostics}
  \vspace{-10pt}
\end{figure}

\paragraph{Training diagnostics.}
We monitor the SAE reconstruction loss throughout both training stages for each target model.
Figure~\ref{fig:sae_training_diagnostics} groups the optimization traces by model family.
Stage 1 losses drop sharply within the first few hundred steps and then settle onto a stable plateau, indicating that the broad-corpus dictionary has converged on OpenWebText2.
At the Stage 2 boundary the loss jumps because the residual-stream distribution shifts to When2Call, and then descends smoothly to a new plateau as the dictionary adapts to tool-use contexts.
All six SAEs follow this pattern, confirming stable optimization across families and scales.

\FloatBarrier
\section{Gating Feature Discovery Across Models}
\label{app:feature_discovery_all_models}

\begin{figure}[h]
  \centering
  \includegraphics[width=0.9\linewidth]{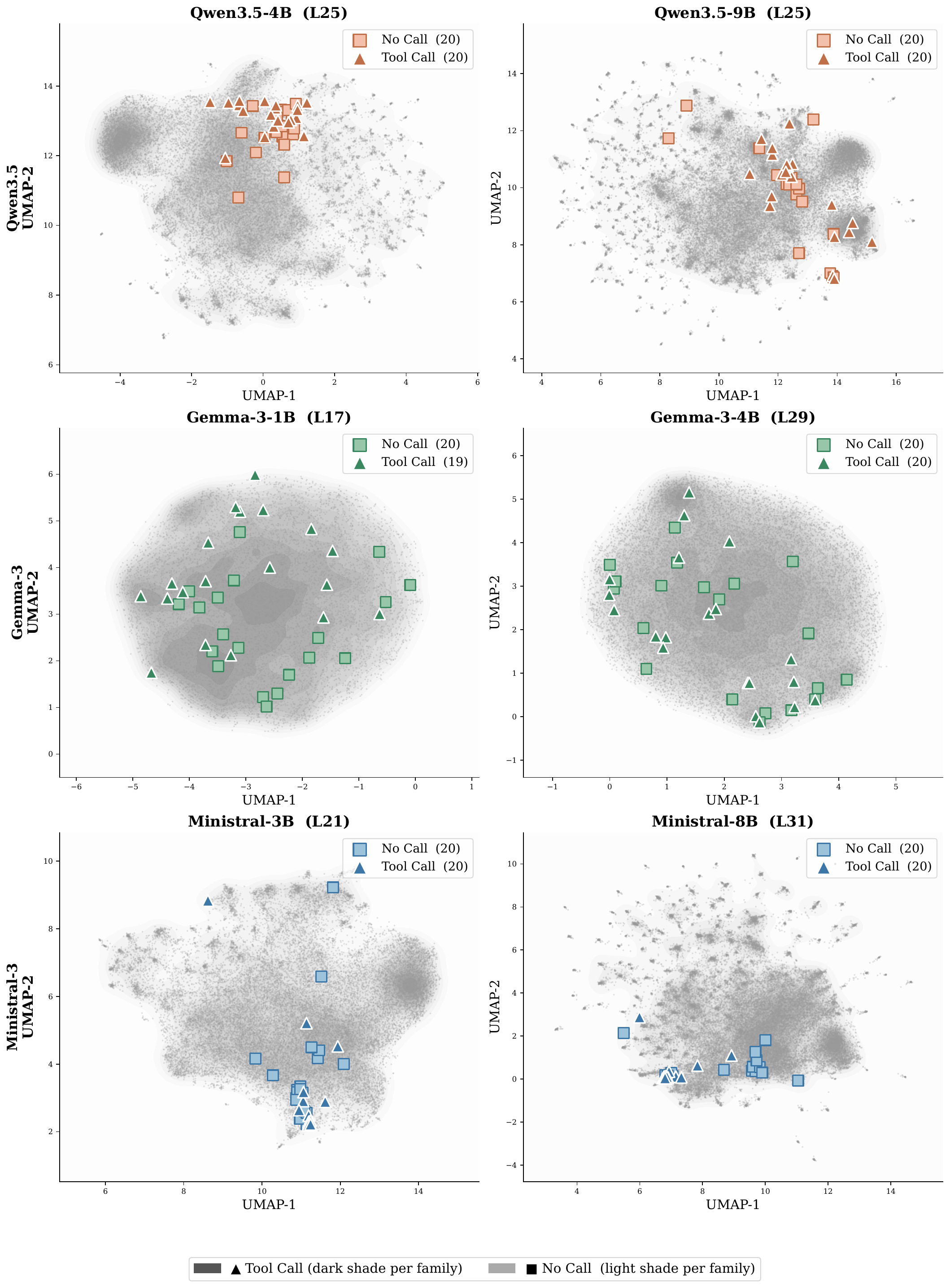}
  \caption{UMAP visualization of the SAE feature dictionary for each target model. Gray points are all SAE features projected to 2D, and the colored markers highlight the top-20 \textsc{tool\_call} (triangles) and \textsc{no\_call} (squares) features returned by the discovery pipeline. Panel titles list the model name and the SAE hook layer.}
  \label{fig:feature_discovery_all_models}
\end{figure}

\begin{figure}[p]
  \centering
  \includegraphics[width=\linewidth]{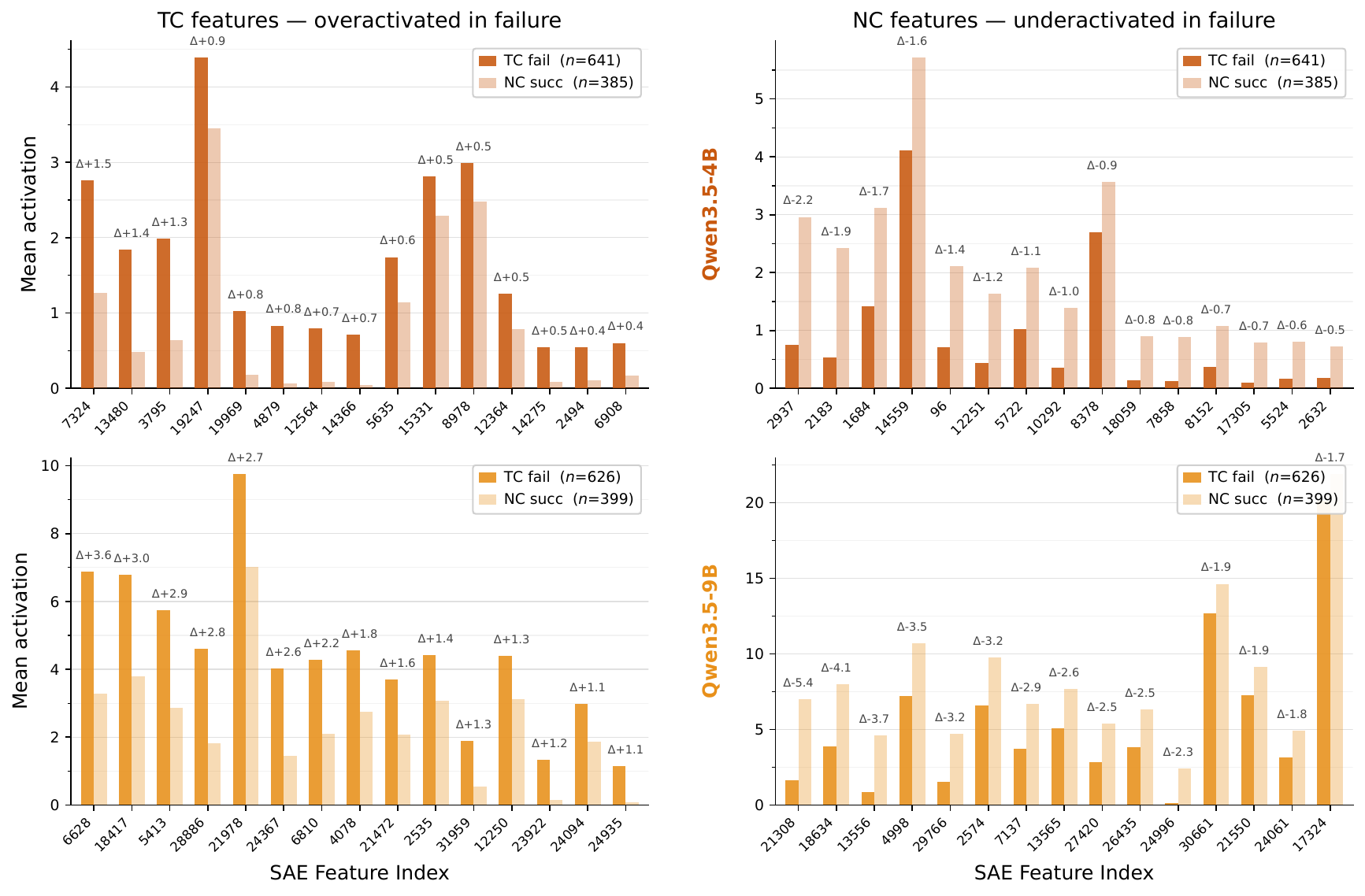}
  \vspace{-4pt}
  \caption{Per-feature mean SAE activation on Tool-Call-failure (over-calling) vs.\ No-Call-success contexts for Qwen3.5-4B (top) and Qwen3.5-9B (bottom), with top-ranked \textsc{tool\_call} features (left) overactivated and \textsc{no\_call} features (right) underactivated under failure.}
  \label{fig:feature_attribution_qwen}
  \vspace{4pt}

  \includegraphics[width=\linewidth]{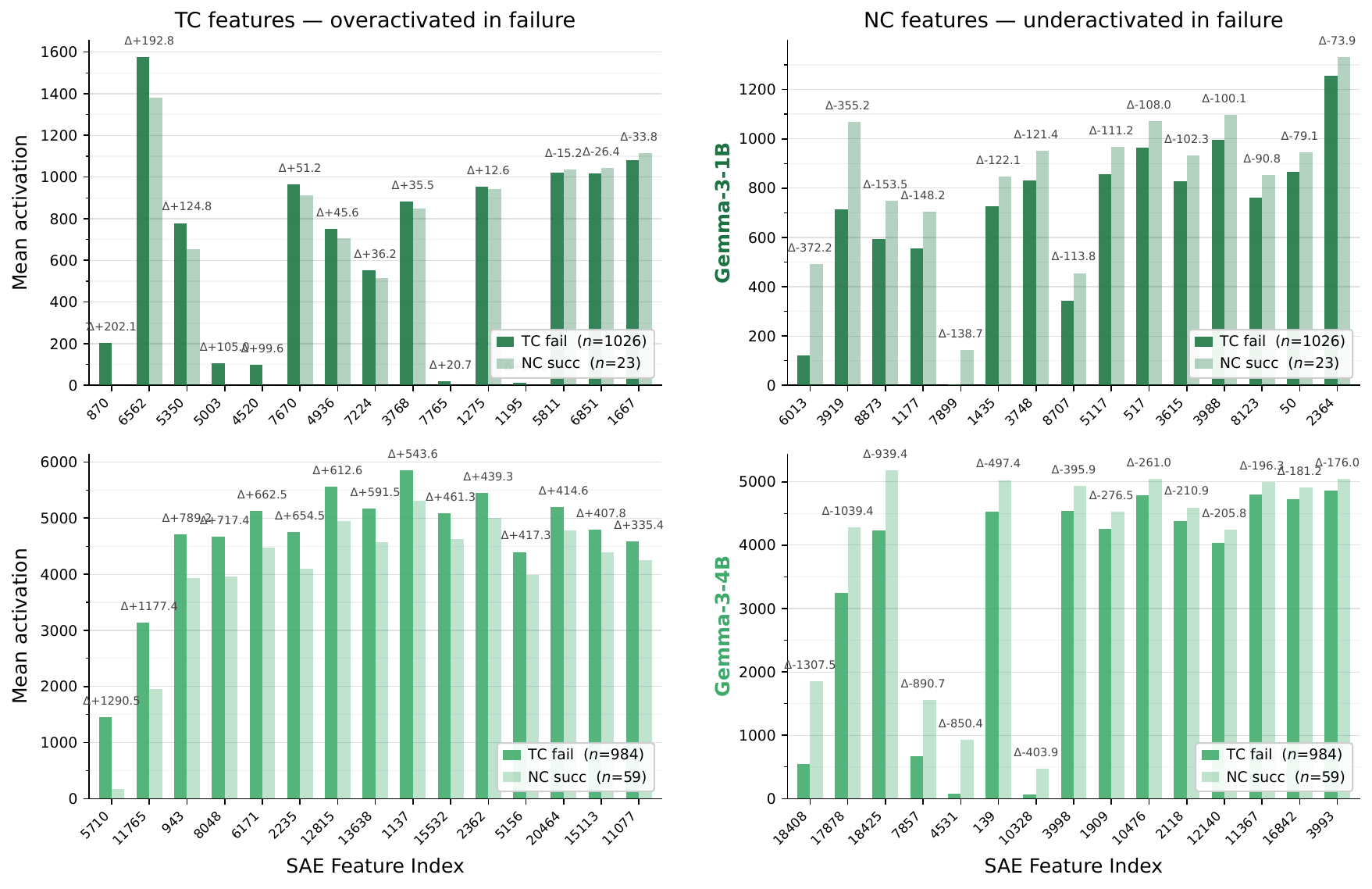}
  \vspace{-4pt}
  \caption{Same as Figure~\ref{fig:feature_attribution_qwen}, for Gemma-3-1B (top) and Gemma-3-4B (bottom).}
  \label{fig:feature_attribution_gemma}
\end{figure}

\begin{figure}[t]
  \centering
  \includegraphics[width=\linewidth]{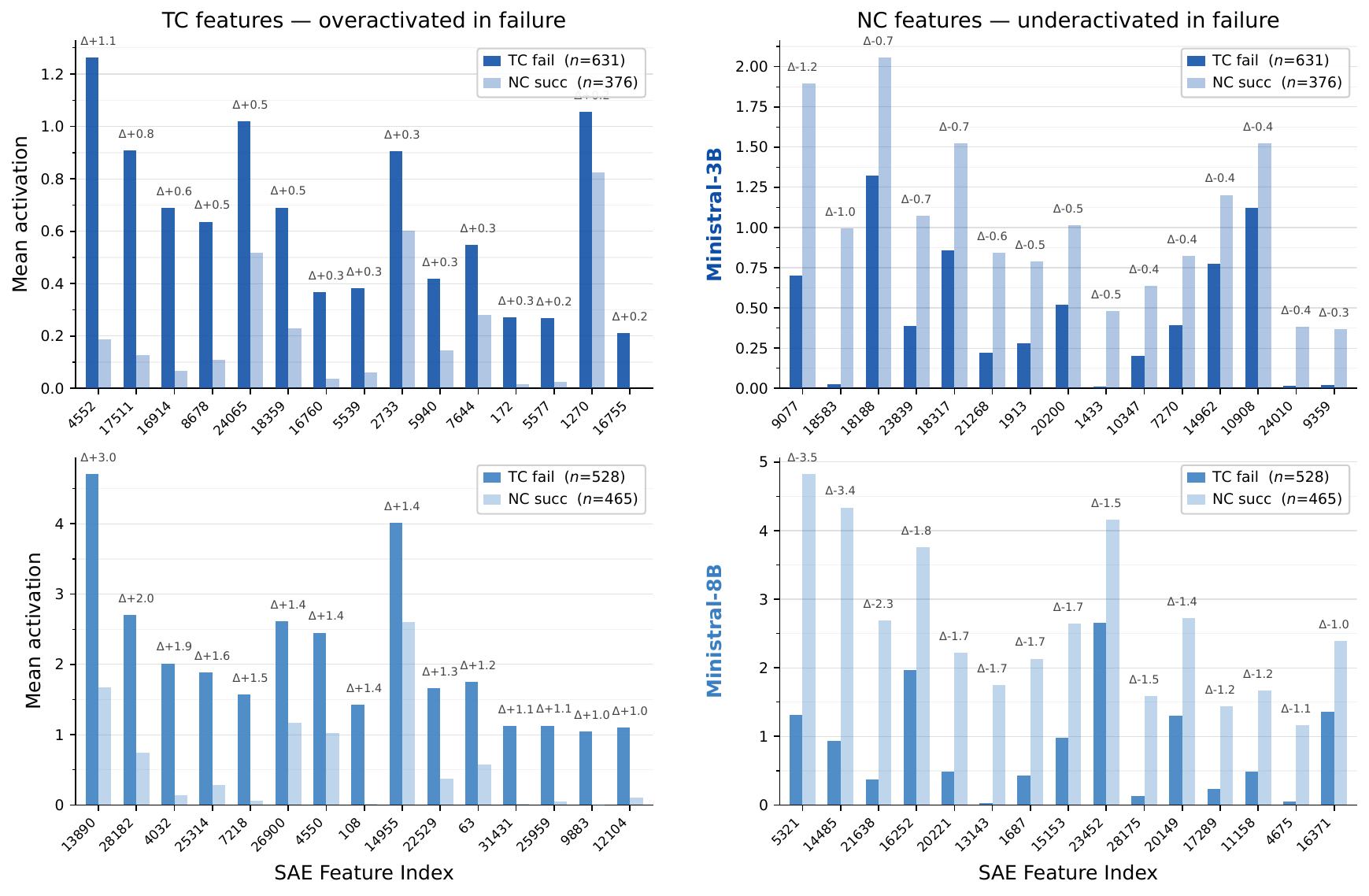}
  \vspace{-4pt}
  \caption{Same as Figure~\ref{fig:feature_attribution_qwen}, for Ministral-3-3B (top) and Ministral-8B (bottom).}
  \label{fig:feature_attribution_ministral}
\end{figure}

Figure~\ref{fig:feature_discovery_all_models} situates the discovered gating features inside each SAE's full feature dictionary.
For the Qwen3.5 and Ministral-3 families, both \textsc{tool\_call} and \textsc{no\_call} features collapse into a single tight cluster on the UMAP manifold, indicating that the call/no-call signal is concentrated in a coherent region of the SAE feature space.
Gemma instead spreads its top features across most of the manifold, so gating information is carried by individually informative but geometrically dispersed directions.
Despite this difference in feature geometry, the discovery pipeline returns a small, behavior-aligned feature set in all three families, consistent with the near-upper-bound probe AUROC reported in \S\ref{sec:validation}.

We further ask how the discovered features behave on failure cases.
Figures~\ref{fig:feature_attribution_qwen}, \ref{fig:feature_attribution_gemma}, and \ref{fig:feature_attribution_ministral} contrast per-feature mean activations between Tool-Call-failure contexts (where the model wrongly issues a tool call) and No-Call-success contexts (where the model correctly withholds the tool call).
Across all six target models, the top-ranked \textsc{tool\_call} features are systematically overactivated under failure and the top-ranked \textsc{no\_call} features are systematically underactivated.
This bidirectional shift is consistent with IBH: failures arise from a coordinated imbalance between the two feature groups rather than from a single group acting in isolation, and the pattern holds even where the UMAP geometry differs across families.

\begin{figure}[!htbp]
  \centering
  \includegraphics[width=1.0\linewidth]{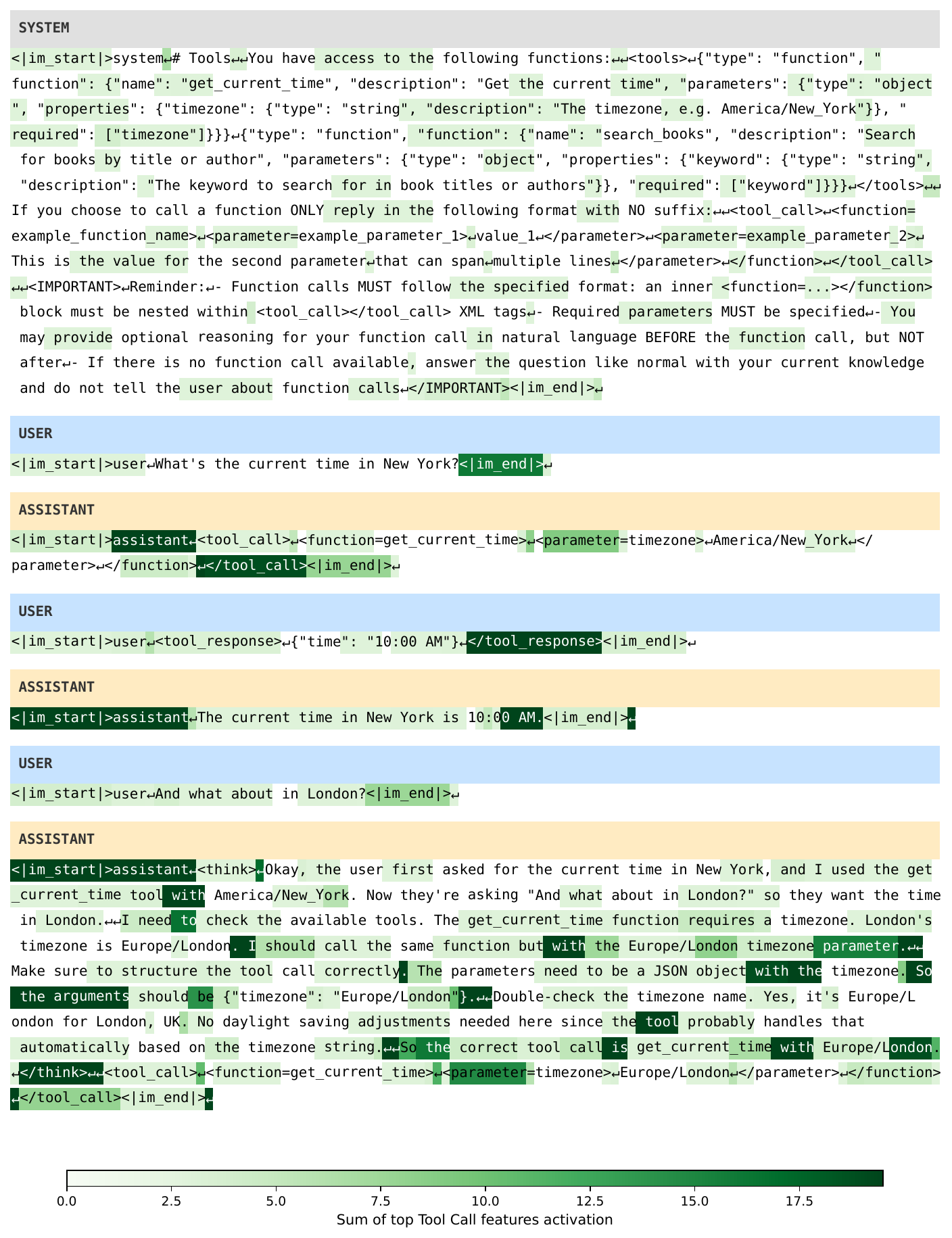}
  \caption{Token-level attribution of the top-10 \textsc{tool\_call} features on a representative context where the model correctly issues a tool call. Green intensity is proportional to each token's contribution to the summed feature activation (x-axis). The strongest signal concentrates in the tool schema definitions and the user query, consistent with these features encoding tool-invocation intent.}
  \label{fig:tc_top10_attribution}
\end{figure}

\begin{figure}[!htbp]
  \centering
  \includegraphics[width=0.88\linewidth]{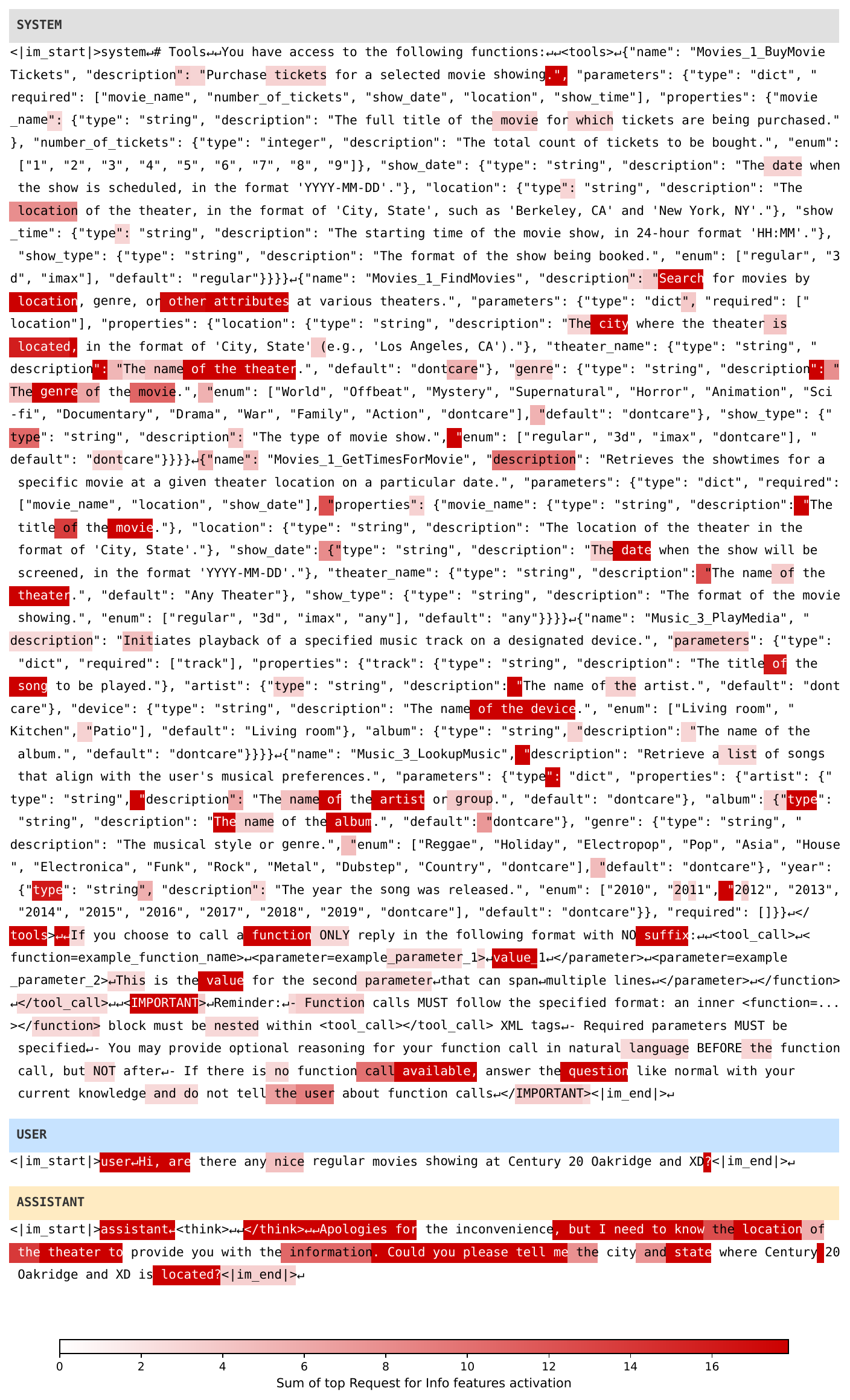}
  \caption{Token-level attribution of the top-10 \textsc{no\_call} features on a representative context where the model correctly withholds a tool call and requests missing information instead. Red intensity is proportional to each token's contribution to the summed feature activation (x-axis). The signal concentrates in the underspecified user query and tool schema, consistent with these features encoding information-insufficiency.}
  \label{fig:rfi_top10_attribution}
\end{figure}

Figures~\ref{fig:tc_top10_attribution} and~\ref{fig:rfi_top10_attribution} trace the top-10 \textsc{tool\_call} and \textsc{no\_call} features back to input tokens on two representative contexts.
Both groups draw most of their activation from the tool schema and the user query.

\section{Adaptive Margin-Calibrated Steering Details}
\label{app:amcs_details}

This section records the calibration and inference steps needed to reproduce the steering vectors, complementing the derivation in \S\ref{sec:amcs}.

\paragraph{Offline calibration.}
For each target model, we cache residual-stream states at the selected SAE hook layer and the corresponding judged model decisions on the calibration set, encode each state with the model-specific SAE, and restrict to examples whose decision $\hat{d}_x \in \{\textsc{call}, \textsc{no\_call}\}$.
On this set, we compute the signed margin $m_r(x)$ over the restricted basis $\mathcal{C}_r \cup \mathcal{N}_r$ following the definition in \S\ref{sec:hypothesis_formalization}, fit Eq.~\eqref{equ:amcs_logistic}, and keep the candidate only when $\beta_r>0$ so that the margin orders \textsc{call} decisions in the expected direction.
We then obtain $\delta_r$ via Eq.~\eqref{equ:amcs_delta}, normalize feature weights as in Eq.~\eqref{equ:amcs_weights}, and form $\mathbf{v}_r$ from Eq.~\eqref{equ:amcs_vector}.
Each candidate budget $r$ induces a distinct feature basis and bias estimate, so calibration is repeated independently for each value of $r$.

\paragraph{Margin-shift approximation.}
Decomposing the contribution of $\mathbf{v}_r$ to the two sides of $m_r$ under a local linear SAE approximation gives $\Delta m_r \approx \alpha\delta_r + (1-\alpha)\delta_r = \delta_r$.
Because decoder directions need not form an orthogonal basis, we treat this as a calibrated target and verify the realized margin shift when reporting steering results.

\paragraph{Algorithm.}
Algorithm~\ref{alg:amcs} summarizes the full offline calibration and online inference procedure.

\begin{algorithm}[t]
  \caption{Adaptive Margin-Calibrated Steering (AMCS)}
  \label{alg:amcs}
  \small
  \begin{algorithmic}[1]
    \Require Cached hidden states $\{\mathbf{h}_x\}_{x \in \mathcal{D}_{\mathrm{cal}}}$, cached model decisions $\{\hat{d}_x\}$, SAE encoder and decoder, hook layer $\ell$, feature rankings $\mathcal{C}$ and $\mathcal{N}$, attribution gaps $\{\Delta^{\mathcal{C}}_j\}$ and $\{\Delta^{\mathcal{N}}_j\}$, feature budget $r$ and allocation coefficient $\alpha$
    \Ensure Steering vector $\mathbf{v}_r$ and steered response $\tilde{y}$
    \Statex
    \State \textbf{Offline calibration}
    \State Select $\mathcal{C}_r \leftarrow \operatorname{Top}_r(\mathcal{C})$ and $\mathcal{N}_r \leftarrow \operatorname{Top}_r(\mathcal{N})$
    \ForAll{$x \in \mathcal{D}_{\mathrm{cal}}$}
      \State $\mathbf{z}_x \leftarrow \operatorname{SAEEnc}(\mathbf{h}_x)$
      \State $m_r(x) \leftarrow \frac{1}{r}\sum_{j \in \mathcal{C}_r}\|\mathbf{d}_j\|_2 z_{x,j} - \frac{1}{r}\sum_{j \in \mathcal{N}_r}\|\mathbf{d}_j\|_2 z_{x,j}$
    \EndFor
    \State $\mathcal{D}_{\mathrm{bin}} \leftarrow \{x \in \mathcal{D}_{\mathrm{cal}}: \hat{d}_x \in \{\textsc{call}, \textsc{no\_call}\}\}$
    \State Fit $\Pr(\hat{d}=1 \mid m_r)=\sigma(\beta_r m_r+\beta_{0,r})$ on $\mathcal{D}_{\mathrm{bin}}$
    \If{$\beta_r \le 0$} \Return \textsc{skip} \EndIf
    \State $\delta_r \leftarrow -\beta_{0,r}/\beta_r$ \Comment{Eq.~\eqref{equ:amcs_delta}}
    \State $\omega^{\mathcal{C}}_j \leftarrow |\Delta^{\mathcal{C}}_j| / \sum_{k \in \mathcal{C}_r}|\Delta^{\mathcal{C}}_k|$ for $j \in \mathcal{C}_r$
    \State $\omega^{\mathcal{N}}_j \leftarrow |\Delta^{\mathcal{N}}_j| / \sum_{k \in \mathcal{N}_r}|\Delta^{\mathcal{N}}_k|$ for $j \in \mathcal{N}_r$
    \State $\mathbf{v}_r \leftarrow \alpha r\delta_r \sum_{j \in \mathcal{C}_r}\omega^{\mathcal{C}}_j\mathbf{d}_j + (1-\alpha)r(-\delta_r)\sum_{j \in \mathcal{N}_r}\omega^{\mathcal{N}}_j\mathbf{d}_j$
    \Statex
    \State \textbf{Online inference}
    \State Register a forward hook at layer $\ell$ that applies $\mathbf{H}_{\ell} \leftarrow \mathbf{H}_{\ell}+\mathbf{v}_r$
    \State Generate $\tilde{y}$ with the hook active
    \State Remove the hook
    \State \Return $\mathbf{v}_r, \tilde{y}$
  \end{algorithmic}
\end{algorithm}

\paragraph{Complexity.}
For a fixed feature budget $r$, constructing $\mathbf{v}_r$ requires a weighted sum of $2r$ decoder columns and costs $O(rd)$.
At inference time, AMCS adds one vector of dimension $d$ to the hooked residual stream.
It introduces no extra model calls, no iterative search over steering coefficients, and no additional learned parameters.

\FloatBarrier
\section{Detailed Transfer and Robustness Results}
\label{app:transfer_details}

\subsection{Cross-Benchmark Transfer}
\label{app:acebench}

\paragraph{Evaluation set.}
The ACEBench set has 100 examples: all 50 \texttt{Special-Incomplete} examples, whose gold decision is \textsc{no\_call}, and 50 randomly sampled \texttt{Normal} single-turn single-function examples, whose gold decision is \textsc{call}.
A correct \textsc{no\_call} response on \texttt{Special-Incomplete} must request the missing information, so direct answers and refusals count as incorrect.

\paragraph{Diagnosis.}
Each model reuses its When2Call SAE and Top-25 \textsc{call}/\textsc{no\_call} features, and encodes the action-boundary activation with the frozen SAE.
Margins are divided by their within-model standard deviation without centering, so $m=0$ remains activation parity.
The diagnostic model is fit on responses judged as \textsc{call} or \textsc{no\_call}, with percentile 95\% intervals from 1,000 example-level bootstrap resamples.
All resamples are identifiable for the four non-Gemma models, and in each of them the full interval for the parity probability lies above $0.5$ and the full interval for $m^{\star}$ lies below zero (Table~\ref{tab:acebench_ibh}).

\begin{table}[!htbp]
  \centering
  \small
  \caption{IBH diagnosis on ACEBench with When2Call features held fixed. $n$ counts responses judged as \textsc{call} or \textsc{no\_call}. Intervals are percentile 95\% bootstrap intervals. Gemma rows are not estimable because \textsc{no\_call} decisions collapse.}
  \label{tab:acebench_ibh}
  \begin{tabular}{lccc}
    \toprule
    Model & $n$ (\textsc{call}/\textsc{no\_call}) & $\Pr(\hat{d}=1\mid m=0)$ & $m^{\star}$ \\
    \midrule
    Qwen3.5-4B & 100 (72/28) & 0.755 [0.648, 0.844] & $-0.438$ [$-0.735$, $-0.209$] \\
    Qwen3.5-9B & 100 (80/20) & 0.990 [0.978, 0.997] & $-2.174$ [$-2.689$, $-1.827$] \\
    Gemma-3-1B & 99 (99/0) & -- & -- \\
    Gemma-3-4B & 100 (98/2) & -- & -- \\
    Ministral-3-3B & 99 (62/37) & 0.744 [0.636, 0.828] & $-0.346$ [$-0.540$, $-0.172$] \\
    Ministral-3-8B & 98 (61/37) & 0.644 [0.527, 0.749] & $-0.174$ [$-0.329$, $-0.033$] \\
    \bottomrule
  \end{tabular}
\end{table}

\paragraph{Steering.}
Directions and strengths are fixed on When2Call and applied unchanged to ACEBench.
CAA uses
\begin{equation}
  v_{\mathrm{CAA}} = \operatorname{mean}(h_{\textsc{no\_call}}) - \operatorname{mean}(h_{\textsc{call}}).
  \label{equ:caa}
\end{equation}
Its coefficient is selected per model from $\{0.2, 0.4, 0.6, 0.8\}$ on When2Call, as the setting with the highest overall accuracy among those that lower call accuracy by at most 5 points.
The selected coefficients are $0.2$, $0.4$, $0.2$, and $0.2$ for Qwen3.5-4B, Qwen3.5-9B, Ministral-3-3B, and Ministral-3-8B.
AMCS uses $r=25$ and $\alpha=0.8$ for all four models.

\FloatBarrier
\subsection{Preservation of Valid Calls}
\label{app:preserved_calls}

We pair the unsteered and AMCS responses on all 1,295 ground-truth tool-call examples in When2Call.
Five models use the fixed Top-25 setting, and Gemma-3-1B uses a reduced Top-6 rerun.
A separate schema-conditioned judge (Appendix~\ref{app:llm_usage}) scores the two responses independently.
Joint accuracy requires an exact function name and argument keys and nested values that match under the tool schema, allowing only non-semantic differences such as JSON key order or \texttt{10} versus \texttt{10.0}.
Unparseable calls are incorrect.
The flip rate is computed where the unsteered model emits a tool call, and joint accuracy on the subset where both responses do (Table~\ref{tab:preserved_calls}).

\begin{table}[!htbp]
  \centering
  \small
  \caption{Effect of AMCS on valid tool calls (When2Call examples whose gold decision is \textsc{call}). Flipped counts unsteered tool calls that AMCS turns into \textsc{no\_call}. Joint accuracy is the rate of exact function name and arguments, on the examples where both the unsteered and the AMCS response emit a tool call. Parentheses give the change in points.}
  \label{tab:preserved_calls}
  \begin{tabular}{lcc}
    \toprule
    Model & Flipped to \textsc{no\_call} & Joint accuracy, unsteered $\to$ AMCS \\
    \midrule
    Qwen3.5-4B & 48/1,238 (3.88\%) & 69.08\% $\to$ 68.99\% ($-0.08$) \\
    Qwen3.5-9B & 86/1,249 (6.89\%) & 73.09\% $\to$ 71.80\% ($-1.29$) \\
    Gemma-3-1B, Top-6 & 301/1,245 (24.18\%) & 24.79\% $\to$ 3.92\% ($-20.87$) \\
    Gemma-3-4B & 4/1,284 (0.31\%) & 54.30\% $\to$ 54.30\% ($0.00$) \\
    Ministral-3-3B & 87/1,182 (7.36\%) & 57.35\% $\to$ 57.63\% ($+0.27$) \\
    Ministral-3-8B & 37/1,184 (3.12\%) & 60.85\% $\to$ 60.59\% ($-0.26$) \\
    \bottomrule
  \end{tabular}
\end{table}

To locate the flips, we rank each model's unsteered tool-call examples by pre-AMCS margin, split them into four equal-count bins, and pool counts across models only after this split (Table~\ref{tab:flip_quartiles}).
Lower margin means weaker \textsc{call}-side activation.
On the five models other than Gemma-3-1B, AMCS preserves 95.73\% of tool calls, 255 of the 262 flips (97.3\%) fall in the lower-margin half, and joint accuracy changes by $-0.27$ points on average.
Gemma-3-1B still flips 8.36\% of tool calls in its highest-margin quartile.
On these five models, AMCS therefore removes mainly the low-margin calls closest to the decision boundary and leaves the content of preserved calls nearly unchanged.
Gemma-3-1B breaks this pattern, so AMCS calls for model-specific downstream checks rather than a universal guarantee that valid calls are preserved.

\begin{table}[!htbp]
  \centering
  \small
  \caption{Tool-call flips by pre-AMCS margin quartile. Q1 is the lowest-margin quartile.}
  \label{tab:flip_quartiles}
  \begin{tabular}{lcc}
    \toprule
    Quartile & Five models other than Gemma-3-1B & Gemma-3-1B, Top-6 \\
    \midrule
    Q1 & 226/1,536 (14.71\%) & 137/312 (43.91\%) \\
    Q2 & 29/1,533 (1.89\%) & 79/311 (25.40\%) \\
    Q3 & 6/1,533 (0.39\%) & 59/311 (18.97\%) \\
    Q4 & 1/1,535 (0.07\%) & 26/311 (8.36\%) \\
    \bottomrule
  \end{tabular}
\end{table}

\FloatBarrier
\subsection{SAE and Margin Ablations}
\label{app:robustness}

\paragraph{SAE retraining.}
For each model we compare the original SAE with five retrained SAEs that change one factor at a time: two hook layers, $K/2$, $2K$, and seed 123 instead of seed 42.
Each new SAE follows the same two-stage training, Top-25 features are rediscovered in it, and the diagnostic model is refit on the decoder-norm-scaled raw-mean margin.
Model outputs and judge labels stay fixed, so differences come only from the SAE.
A setting supports IBH when $\beta>0$, $\Pr(\hat{d}=1\mid m=0)>0.5$, and $m^{\star}<0$.
Raw $m^{\star}$ is not compared across settings, because independently trained SAEs differ in activation scale.

Thirty-four of the 36 settings support IBH (Tables~\ref{tab:sae_ablation_qwen} and~\ref{tab:sae_ablation_all}).
The two exceptions, Ministral-3-3B at Layer 24 and Ministral-3-8B at $K/2$, have a slightly positive $m^{\star}$ but positive slopes, $5.024$ and $3.902$.
Their margins still agree with the emitted decision on 95.7\% and 97.0\% of examples when \textsc{call} activation is higher, and on 92.0\% and 94.5\% when \textsc{no\_call} activation is higher.

\begin{table}[t]
  \centering
  \small
  \caption{IBH under SAE retraining for the five models not shown in Table~\ref{tab:sae_ablation_qwen}. Each model has the original setting and five single-factor variants. Together with Qwen3.5-4B, these are the 36 settings summarized in Section~\ref{sec:robustness}.}
  \label{tab:sae_ablation_all}
  \setlength{\tabcolsep}{4pt}
  \begin{tabular}{llrrrrcc}
    \toprule
    Model & Setting & Layer & $K$ & Seed & $\beta$ & $\Pr(\hat{d}=1\mid m=0)$ & $m^{\star}$ \\
    \midrule
    \multirow{6}{*}{Qwen3.5-9B} & Original & 25 & 128 & 42 & 1.338 & 0.996 & $-4.177$ \\
     & Layer 22 & 22 & 128 & 42 & 2.187 & 0.939 & $-1.247$ \\
     & Layer 28 & 28 & 128 & 42 & 1.083 & 0.865 & $-1.717$ \\
     & $K/2$ & 25 & 64 & 42 & 1.394 & 0.945 & $-2.046$ \\
     & $2K$ & 25 & 256 & 42 & 1.377 & 0.988 & $-3.211$ \\
     & Seed 123 & 25 & 128 & 123 & 1.298 & 0.966 & $-2.584$ \\
    \midrule
    \multirow{6}{*}{Gemma-3-1B} & Original & 17 & 36 & 42 & 0.009 & 0.991 & $-496.909$ \\
     & Layer 14 & 14 & 36 & 42 & 0.008 & 1.000 & $-950.271$ \\
     & Layer 20 & 20 & 36 & 42 & 0.006 & 0.997 & $-904.578$ \\
     & $K/2$ & 17 & 18 & 42 & 0.012 & 0.988 & $-368.191$ \\
     & $2K$ & 17 & 72 & 42 & 0.017 & 0.873 & $-116.301$ \\
     & Seed 123 & 17 & 36 & 123 & 0.009 & 0.997 & $-604.008$ \\
    \midrule
    \multirow{6}{*}{Gemma-3-4B} & Original & 29 & 80 & 42 & 0.008 & 0.860 & $-241.619$ \\
     & Layer 26 & 26 & 80 & 42 & 0.003 & 0.993 & $-1439.793$ \\
     & Layer 32 & 32 & 80 & 42 & 0.003 & 0.996 & $-1990.850$ \\
     & $K/2$ & 29 & 40 & 42 & 0.002 & 1.000 & $-4760.214$ \\
     & $2K$ & 29 & 160 & 42 & 0.006 & 0.964 & $-562.989$ \\
     & Seed 123 & 29 & 80 & 123 & 0.008 & 0.957 & $-392.407$ \\
    \midrule
    \multirow{6}{*}{Ministral-3-3B} & Original & 21 & 96 & 42 & 9.403 & 0.927 & $-0.270$ \\
     & Layer 18 & 18 & 96 & 42 & 13.926 & 0.890 & $-0.150$ \\
     & Layer 24 & 24 & 96 & 42 & 5.024 & 0.422 & $0.063$ \\
     & $K/2$ & 21 & 48 & 42 & 7.968 & 0.663 & $-0.085$ \\
     & $2K$ & 21 & 192 & 42 & 9.135 & 0.913 & $-0.257$ \\
     & Seed 123 & 21 & 96 & 123 & 8.888 & 0.928 & $-0.288$ \\
    \midrule
    \multirow{6}{*}{Ministral-3-8B} & Original & 31 & 128 & 42 & 3.988 & 0.840 & $-0.415$ \\
     & Layer 28 & 28 & 128 & 42 & 4.461 & 0.913 & $-0.528$ \\
     & Layer 24 & 24 & 128 & 42 & 7.562 & 0.846 & $-0.226$ \\
     & $K/2$ & 31 & 64 & 42 & 3.902 & 0.433 & $0.069$ \\
     & $2K$ & 31 & 256 & 42 & 4.059 & 0.612 & $-0.113$ \\
     & Seed 123 & 31 & 128 & 123 & 3.606 & 0.852 & $-0.486$ \\
    \bottomrule
  \end{tabular}%
\end{table}

\paragraph{Feature budget.}
On the original SAE, each side uses its top-$N$ features for $N\in\{5,10,15,20,25,30\}$, with the raw mean-difference margin.
Gemma-3-1B has only 19 \textsc{call} features, so its Top-20, Top-25, and Top-30 settings use 19 \textsc{call} features against 20, 25, and 30 \textsc{no\_call} features.
IBH holds in 35 of the 36 settings (Table~\ref{tab:topk_ablation}).
The only miss is Gemma-3-4B at Top-5, where $\Pr(\hat{d}=1\mid m=0)=0.245$ and $m^{\star}=+517.893$.
The Qwen3.5-4B Top-25 row lists $\beta=4.008$, against $4.006$ in Table~\ref{tab:sae_ablation_qwen} for the same SAE.

\begin{table}[!htbp]
  \centering
  \small
  \caption{Feature-budget ablation on the original SAE. Each side uses its top-$N$ features, except Gemma-3-1B, whose \textsc{call} set contains only 19 features. The margin is the raw mean difference.}
  \label{tab:topk_ablation}
  \setlength{\tabcolsep}{4pt}
  \begin{tabular}{llrccc}
    \toprule
    Model & $N$ & Features & $\beta$ & $\Pr(\hat{d}=1\mid m=0)$ & $m^{\star}$ \\
    \midrule
    \multirow{6}{*}{Qwen3.5-4B} & 5 & 5/5 & 1.694 & 0.915 & $-1.402$ \\
     & 10 & 10/10 & 2.444 & 0.945 & $-1.166$ \\
     & 15 & 15/15 & 2.938 & 0.876 & $-0.664$ \\
     & 20 & 20/20 & 3.518 & 0.894 & $-0.605$ \\
     & 25 & 25/25 & 4.008 & 0.894 & $-0.533$ \\
     & 30 & 30/30 & 4.527 & 0.887 & $-0.454$ \\
    \midrule
    \multirow{6}{*}{Qwen3.5-9B} & 5 & 5/5 & 0.818 & 0.896 & $-2.628$ \\
     & 10 & 10/10 & 0.878 & 0.961 & $-3.664$ \\
     & 15 & 15/15 & 0.940 & 0.978 & $-4.018$ \\
     & 20 & 20/20 & 1.117 & 0.997 & $-5.354$ \\
     & 25 & 25/25 & 1.338 & 0.996 & $-4.177$ \\
     & 30 & 30/30 & 1.471 & 0.996 & $-3.694$ \\
    \midrule
    \multirow{6}{*}{Gemma-3-1B} & 5 & 5/5 & 0.005 & 0.926 & $-501.378$ \\
     & 10 & 10/10 & 0.007 & 0.974 & $-525.545$ \\
     & 15 & 15/15 & 0.007 & 0.987 & $-606.731$ \\
     & 20 & 19/20 & 0.009 & 0.992 & $-529.477$ \\
     & 25 & 19/25 & 0.009 & 0.991 & $-496.909$ \\
     & 30 & 19/30 & 0.010 & 0.985 & $-433.546$ \\
    \midrule
    \multirow{6}{*}{Gemma-3-4B} & 5 & 5/5 & 0.002 & 0.245 & $+517.893$ \\
     & 10 & 10/10 & 0.004 & 0.685 & $-215.102$ \\
     & 15 & 15/15 & 0.005 & 0.903 & $-430.247$ \\
     & 20 & 20/20 & 0.006 & 0.891 & $-326.064$ \\
     & 25 & 25/25 & 0.008 & 0.860 & $-241.619$ \\
     & 30 & 30/30 & 0.009 & 0.985 & $-471.839$ \\
    \midrule
    \multirow{6}{*}{Ministral-3-3B} & 5 & 5/5 & 3.985 & 0.930 & $-0.648$ \\
     & 10 & 10/10 & 6.407 & 0.921 & $-0.383$ \\
     & 15 & 15/15 & 7.209 & 0.929 & $-0.357$ \\
     & 20 & 20/20 & 8.294 & 0.903 & $-0.268$ \\
     & 25 & 25/25 & 9.403 & 0.927 & $-0.270$ \\
     & 30 & 30/30 & 10.186 & 0.907 & $-0.224$ \\
    \midrule
    \multirow{6}{*}{Ministral-3-8B} & 5 & 5/5 & 2.316 & 0.548 & $-0.084$ \\
     & 10 & 10/10 & 2.804 & 0.746 & $-0.385$ \\
     & 15 & 15/15 & 3.210 & 0.797 & $-0.426$ \\
     & 20 & 20/20 & 3.539 & 0.690 & $-0.225$ \\
     & 25 & 25/25 & 3.988 & 0.840 & $-0.415$ \\
     & 30 & 30/30 & 4.053 & 0.833 & $-0.396$ \\
    \bottomrule
  \end{tabular}
\end{table}

\paragraph{Margin construction.}
At Top-25 we compare three margins: the raw mean difference, a feature-standardized difference that divides each feature by its evaluation-set standard deviation, and the difference of the two side means divided by their sum.
All 18 settings support IBH (Table~\ref{tab:margin_ablation}).
Raw $m^{\star}$ is not compared across constructions, because the units differ.
None of the margins is mean-centered.
For a centered margin $m'=m-c$, the fit becomes $\beta m+\beta_0=\beta m'+(\beta_0+\beta c)$, so $m'=0$ marks the dataset mean rather than activation parity and the intercept moves with the arbitrary choice of $c$.
A centered fit would no longer test the parity claim, so the ablations vary scale and feature weighting while keeping $m=0$ at parity.

\begin{table}[!htbp]
  \centering
  \small
  \caption{Margin-construction ablation at Top-25 on the original SAE. Standardized divides each feature by its evaluation-set standard deviation and does not subtract the mean. Difference/sum divides the difference of the two side means by their sum.}
  \label{tab:margin_ablation}
  \setlength{\tabcolsep}{4pt}
  \begin{tabular}{llrccc}
    \toprule
    Model & Margin & Features & $\beta$ & $\Pr(\hat{d}=1\mid m=0)$ & $m^{\star}$ \\
    \midrule
    \multirow{3}{*}{Qwen3.5-4B} & Raw & 25/25 & 4.008 & 0.894 & $-0.533$ \\
     & Standardized & 25/25 & 3.019 & 0.782 & $-0.424$ \\
     & Difference/sum & 25/25 & 8.569 & 0.868 & $-0.220$ \\
    \midrule
    \multirow{3}{*}{Qwen3.5-9B} & Raw & 25/25 & 1.338 & 0.996 & $-4.177$ \\
     & Standardized & 25/25 & 3.368 & 0.959 & $-0.936$ \\
     & Difference/sum & 25/25 & 10.285 & 0.990 & $-0.450$ \\
    \midrule
    \multirow{3}{*}{Gemma-3-1B} & Raw & 19/25 & 0.009 & 0.991 & $-496.909$ \\
     & Standardized & 19/25 & 1.374 & 0.994 & $-3.708$ \\
     & Difference/sum & 19/25 & 2.453 & 0.988 & $-1.798$ \\
    \midrule
    \multirow{3}{*}{Gemma-3-4B} & Raw & 25/25 & 0.008 & 0.860 & $-241.619$ \\
     & Standardized & 25/25 & 3.437 & 0.946 & $-0.833$ \\
     & Difference/sum & 25/25 & 6.114 & 0.958 & $-0.510$ \\
    \midrule
    \multirow{3}{*}{Ministral-3-3B} & Raw & 25/25 & 9.403 & 0.927 & $-0.270$ \\
     & Standardized & 25/25 & 4.853 & 0.921 & $-0.507$ \\
     & Difference/sum & 25/25 & 9.741 & 0.924 & $-0.256$ \\
    \midrule
    \multirow{3}{*}{Ministral-3-8B} & Raw & 25/25 & 3.988 & 0.840 & $-0.415$ \\
     & Standardized & 25/25 & 4.135 & 0.714 & $-0.221$ \\
     & Difference/sum & 25/25 & 8.216 & 0.809 & $-0.176$ \\
    \bottomrule
  \end{tabular}
\end{table}

\section{Scaling, Pretraining, and Reasoning}
\label{app:additional_settings}

\subsection{Models above 10B}
\label{app:larger_models}

The main experiments use models below 10B parameters.
We train separate SAEs for Qwen3-14B (14.8B parameters, Top-25 features per side) and Gemma-3-12B-IT (12.0B parameters, Top-10 features per side) and rediscover the \textsc{call} and \textsc{no\_call} features on the full 3,652-example When2Call test set.
Qwen3-14B is a Qwen3 checkpoint, not Qwen3.5, and neither model is evaluated with AMCS.
Margins are divided by the within-model standard deviation without centering, so $m=0$ remains activation parity.

\begin{table}[!htbp]
  \centering
  \small
  \caption{IBH on two models above 10B. $n$ counts responses judged as \textsc{call} or \textsc{no\_call}. $m^{\star}$ is in within-model standard-deviation units.}
  \label{tab:larger_models}
  \begin{tabular}{lccc}
    \toprule
    Model & $n$ (\textsc{call}/\textsc{no\_call}) & $\Pr(\hat{d}=1\mid m=0)$ & $m^{\star}$ \\
    \midrule
    Qwen3-14B & 2,297 (2,060/237) & 0.996 & $-0.856$ \\
    Gemma-3-12B-IT & 2,317 (1,976/341) & 0.998 & $-1.615$ \\
    \bottomrule
  \end{tabular}
\end{table}

Both models have positive slopes, parity probabilities above $0.5$, and negative neutral boundaries (Table~\ref{tab:larger_models}).
For Gemma-3-12B-IT, 992 examples lie at $m<0$ and 1,325 at $m>0$, so parity lies inside the observed margin range.
Each SAE is trained independently, so the normalized margins are not a shared coordinate, and these two models do not establish a relation between parameter count and offset size.

\FloatBarrier
\subsection{Pretrained Checkpoint}
\label{app:base_checkpoint}

We compare Qwen3.5-4B with the officially released pretrained-only Qwen3.5-4B-Base.
The base weights stay frozen, and a separate SAE is trained on it only as a measurement model, with the same nominal layer, dictionary size, sparsity, feature discovery, prompts, judge, and tool-aware first-turn format as Qwen3.5-4B.
The two SAE bases differ, so we compare the direction of the IBH criteria rather than the size of the offsets.

\begin{table}[!htbp]
  \centering
  \small
  \caption{IBH on Qwen3.5-4B-Base and the instruction-tuned Qwen3.5-4B. $n$ counts responses judged as \textsc{call} or \textsc{no\_call}. Intervals are 95\% confidence intervals.}
  \label{tab:base_ibh}
  \begin{tabular}{lccc}
    \toprule
    Checkpoint & $n$ (\textsc{call}/\textsc{no\_call}) & $\Pr(\hat{d}=1\mid m=0)$ & $m^{\star}$ \\
    \midrule
    Qwen3.5-4B-Base & 2,345 (1,947/398) & 0.9621 [0.9509, 0.9730] & $-0.691$ [$-0.728$, $-0.651$] \\
    Qwen3.5-4B & 2,308 (1,889/419) & 0.8943 [0.8746, 0.9149] & $-0.533$ [$-0.588$, $-0.487$] \\
    \bottomrule
  \end{tabular}
\end{table}

Both slopes are positive.
For the base checkpoint, the full interval for the parity probability lies above $0.5$ and the full interval for $m^{\star}$ lies below zero (Table~\ref{tab:base_ibh}), so the offset is already detectable before the publicly documented post-training stage.

Behaviorally, on the same 3,652 prompts, post-training lowers the overall call rate from 65.85\% to 59.23\% ($-6.63$ points) and the false-call rate on ground-truth \textsc{no\_call} prompts from 64.22\% to 60.36\% ($-3.86$ points, $p=0.0051$).
The correct \textsc{no\_call} rate rises only from 35.03\% to 36.25\% ($+1.22$ points, $p=0.383$).
The $p$-values are paired exact tests on the same prompts.
The two rates are not complements, because the model may also answer directly or state that it cannot answer.
Post-training thus reduces the propensity to call but does not create the offset from scratch, and the comparison does not identify which training stage produces it.

\FloatBarrier
\subsection{Reasoning On and Off}
\label{app:reasoning}

We compare Qwen3.5-4B with reasoning off and on over 2,357 ground-truth \textsc{call}/\textsc{no\_call} examples, using the frozen Stage-2 SAE at Layer 25, Top-25 features, greedy decoding, and the same judge.
The judge sees only the final answer after \texttt{</think>}.
The residual state is read after \texttt{</think>} for reasoning-on, and at the action boundary after an empty reasoning block for reasoning-off.
The reasoning-off responses were regenerated for this comparison.
Their IBH estimates match Section~\ref{sec:diagnosis}, while their behavioral accuracies differ slightly from Table~\ref{tab:steering_validation}.

\begin{table}[!htbp]
  \centering
  \small
  \caption{Qwen3.5-4B with reasoning off and on. Call and no-call columns are accuracies (\%).}
  \label{tab:reasoning}
  \begin{tabular}{lcccc}
    \toprule
    Setting & Call & No-call & $\Pr(\hat{d}=1\mid m=0)$ & $m^{\star}$ \\
    \midrule
    Reasoning-off & 96.37 & 36.25 & 0.894 & $-0.533$ \\
    Reasoning-on & 95.52 & 59.32 & 0.934 & $-0.573$ \\
    On $-$ off & $-0.85$ & $+23.07$ & $+0.0395$ & $-0.040$ \\
    \bottomrule
  \end{tabular}
\end{table}

Reasoning raises no-call accuracy by 23.07 points with call accuracy nearly unchanged, but the call preference at parity remains (Table~\ref{tab:reasoning}).
Under 2,000 source-cluster paired bootstrap resamples, the parity probability rises by $+0.0395$ (95\% interval $[+0.0127,+0.0643]$).
The change in $m^{\star}$, $-0.040$ with interval $[-0.109,+0.037]$, crosses zero, so we do not claim that reasoning moves the neutral boundary.
The mean margin barely moves on ground-truth \textsc{call} examples ($+0.734$ to $+0.726$) but falls from $-0.165$ to $-0.584$ on ground-truth \textsc{no\_call} examples.
The no-call mean thus moves from the call side of the off-condition boundary ($-0.533$) to the no-call side of the on-condition boundary ($-0.573$).
Reasoning strengthens instance-level evidence for missing information without removing the call bias at activation parity.

\FloatBarrier
\section{LLM Usage}
\label{app:llm_usage}

This work studies LLMs as target models for tool-use decisions.
We also use an independent LLM judge to classify generated responses into \textsc{call} and \textsc{no\_call} behavior labels for feature discovery, calibration, and evaluation.
The judge prompt is reported in Appendix~\ref{app:judge_prompt}.
The same judging procedure is used for baseline and steered conditions.
A separate judge, Qwen3.6-27B with deterministic decoding, scores function names and arguments for the preserved-call analysis in Appendix~\ref{app:preserved_calls}.
It is not the four-way behavior judge above.
AMCS itself is defined by SAE activations, a fitted margin model, and a fixed hook, rather than by an LLM component.

\section{Perplexity of Steered Outputs}
\label{app:perplexity}

\begin{table}[h]
  \centering
  \small
  \caption{Next-token perplexity of steered outputs on the When2Call test split. Baseline is the unmodified model. Suppress and Promote are one-sided ablations using only the call-suppression or no-call-promotion component of the AMCS vector, respectively. AMCS is the full two-sided intervention. The parameter $\alpha \in \{0.2, 0.6, 1.0\}$ controls the allocation between the two sides. Extreme values for Gemma3 reflect the weak \textsc{call}/\textsc{no\_call} feature separation noted in \S\ref{sec:steering_validation}.}
  \label{tab:perplexity}
  \setlength{\tabcolsep}{4pt}
  \resizebox{\textwidth}{!}{%
  \begin{tabular}{llrrrrrr}
    \toprule
    Method & $\alpha$ & Qwen3.5-4B & Qwen3.5-9B & Gemma3-1B & Gemma3-4B & Ministral3-3B & Ministral3-8B \\
    \midrule
    Baseline & -- & 1.32 & 1.37 & 1.04 & 1.02 & 3.17 & 2.93 \\
    \midrule
    \multirow{3}{*}{Suppress} & 0.2 & 1.33 & 1.37 & 1.25 & 1.06 & 3.12 & 2.85 \\
     & 0.6 & 1.36 & 1.39 & $3.19\times10^{7}$ & 1.33 & 2.93 & 2.58 \\
     & 1.0 & 1.45 & 1.42 & 225.02 & 57317.43 & 2.12 & 1.63 \\
    \midrule
    \multirow{3}{*}{Promote} & 0.2 & 1.34 & 1.36 & 1.13 & 1.03 & 3.14 & 2.90 \\
     & 0.6 & 1.37 & 1.35 & 2.38 & 1.20 & 3.09 & 2.84 \\
     & 1.0 & 1.43 & 1.37 & 7.44 & 1.54 & 3.02 & 2.80 \\
    \midrule
    \multirow{3}{*}{AMCS} & 0.2 & 1.36 & 1.38 & 2.29 & 1.02 & 3.08 & 2.90 \\
     & 0.6 & 1.36 & 1.42 & 1.53 & 1.02 & 3.05 & 2.84 \\
     & 1.0 & 1.38 & 1.46 & 392.70 & 1.02 & 3.01 & 2.80 \\
    \bottomrule
  \end{tabular}%
  }
\end{table}

Note that models use different prompt templates and produce responses of different types (\textsc{tool\_call} vs.\ \textsc{no\_call}), so perplexity values are only meaningful within a single model compared against its own baseline, not across models or response types.
For Qwen3.5 and Ministral3, all three methods maintain perplexity within a small margin of the unsteered baseline across all tested $\alpha$ values, confirming that residual-stream additions do not destabilize generation.
Gemma3 is the exception: Suppress at higher $\alpha$ produces extreme values ($3.19\times10^7$ for Gemma3-1B at $\alpha=0.6$, $57317.43$ for Gemma3-4B at $\alpha=1.0$), and AMCS at $\alpha=1.0$ also degrades for the 1B model ($392.70$).
This instability mirrors the weak \textsc{call}/\textsc{no\_call} feature separation already noted for Gemma3 in \S\ref{sec:steering_validation}, where Gemma-3-4B is a null case and Gemma-3-1B is a failure case for AMCS.

\section{Broader Impacts}
\label{app:broader_impacts}

This work can improve the reliability and efficiency of tool-using agents by reducing unnecessary tool calls, especially in cases where the model should ask for missing information instead of calling a tool.
Such behavior can lower API cost, reduce avoidable external actions, and make agent decisions easier to audit through feature-level diagnostics.

At the same time, steering a model's tool-use propensity can introduce deployment risks if applied without task-specific validation.
Overly conservative steering may suppress necessary tool calls, while overly aggressive steering may amplify automation errors.
We therefore view AMCS as a diagnostic and mitigation tool that should be evaluated with call accuracy, no-call accuracy, and downstream task checks before deployment.

\section{Experiments Compute Resources}
\label{app:compute_resources}

All experiments were conducted on NVIDIA A800 (80GB) GPUs. 
For individual experiments, SAE training on a single model requires approximately 4 GPU-hours, while \textsc{When2Call} evaluation on a single model takes approximately 3 GPU-hours. Experiments involving activation steering require approximately 4 GPU-hours per model. 
The main results in Table~\ref{tab:steering_validation} encompass 98 \textsc{When2Call} evaluation runs, amounting to roughly $4 \times 98 = 392$ GPU-hours in total, which were completed in approximately 2 days using 8 GPUs in parallel. 
In aggregate, all experiments reported in this paper were conducted over a period of two months with dedicated access to 8 GPUs.

\end{document}